\documentclass[runningheads]{llncs}
\usepackage{graphicx}
\usepackage{bm}
\usepackage{tikz}
\usepackage{pgfplots}
\usepackage{float}
\usepackage{multirow}
\usepackage{xcolor}
\usetikzlibrary{positioning,calc}
\usepackage{makecell}
\usepackage{tabularx}
\usepackage{booktabs}

\pgfplotsset{compat=1.14}
\begin{document}

\title{Transfer Learning with Sparse Associative Memories}
\titlerunning{Transfer Learning with Sparse Associatie Memories}

\author{Quentin Jodelet\inst{1,2} \and
Vincent Gripon\inst{1} \and
Masafumi Hagiwara\inst{2}}
\authorrunning{Q. Jodelet et al.}

\institute{IMT Atlantique, Technopole Brest Iroise, 29238 Brest, France \and
Keio University, Yagami Campus, 223-8522 Yokohama, Japan}

\maketitle              
\begin{abstract}
\let\thefootnote\relax\footnotetext{Presented at the 28th International Conference on Artificial Neural Networks.}
  In this paper, we introduce a novel layer designed to be used as the output of pre-trained neural networks in the context of classification. Based on Associative Memories, this layer can help design deep neural networks which support incremental learning and that can be (partially) trained in real time on embedded devices. Experiments on the ImageNet dataset and other different domain specific datasets show that it is possible to design more flexible and faster-to-train Neural Networks at the cost of a slight decrease in accuracy.

\end{abstract}
\section{Introduction}
During the past decade, deep neural networks, and more specifically Deep Convolutional Neural Networks have been established as the state-of-the-art solution for various problems of Computer Vision such as image classification~\cite{NIPS2012_4824,simonyan2014very,he2016deep,szegedy2016rethinking}, image segmentation \cite{long2015fully,he2017mask,chen2017rethinking} and object tracking~\cite{7410714,bertinetto2016fully}.

A standard Deep Neural Network relies on millions of trained parameters and thus requires millions of floating point operations in order to compute the output corresponding to a given input. Consequently, the use of deep neural networks for inference in real time tasks requires massive computing power and large amounts of memory. However embedded devices have important limitations in terms of computing power, memory and battery usage, so that deep neural networks are difficult to implement. Many research works have been carried out in order to produce faster deep neural networks architectures at run-time ; new specific architectures have been developed specifically for real-time execution~\cite{redmon2016you} and embedded systems~\cite{howard2017mobilenets,sandler2018mobilenetv2}. Quantization~\cite{7780890,zhou2017incremental,han2015deep,han2015learning}, and more generally binarization~\cite{courbariaux2016binarized,rastegari2016xnor,zhou2016dorefa}, of deep neural networks is the preferred solution for running inference on embedded devices. It permits, at the cost of a little decrease of the accuracy, to replace the computationally intensive floating-point operations by low-bit operations which can be more efficiently implemented, especially on FPGA.

Nevertheless, those research are mostly focused on the efficient execution of the inference on the embedded device and not on the even more complex training procedure which requires going through a large dataset multiple time. As of today, this procedure is generally performed offline using specific hardware such as GPUs or TPUs. 
Moreover, neural networks trained sequentially using backpropagation algorithm have a high propensity to steeply forget previous tasks when learning new ones regardless of whether they are used for reinforcement learning or supervised learning. This situation, referred to as catastrophic forgetting~\cite{french1999catastrophic,goodfellow2013empirical}, happens because the weights of the network previously optimized for the first tasks are overwritten in order to correctly achieve the new task. Thus, adding new elements to the dataset (or new classes) can be handled only by restarting the training from scratch. 

In order to benefit from the accuracy of deep neural networks without having to train them, a common solution is to rely on transfer learning~\cite{Pan:2010:STL:1850483.1850545}. In the context of computer vision, transfer learning consists in using deep neural networks pre-trained on a large dataset such as ImageNet~\cite{deng2009imagenet}, Microsoft COCO~\cite{lin2014microsoft} or Google OpenImages~\cite{OpenImages}, in order to obtain a generic image representation for other tasks. It is then possible to address new classification tasks on a different dataset by using pre-trained models as feature extractors, which may then be fine-tuned and combined with a simple, sometimes incremental, classifier~\cite{donahue2014decaf,sharif2014cnn,yosinski2014transferable,azizpour2016factors}

In this paper, by using Self-Organizing Maps and Sparse Associative Memories, we propose a new Neural Network model meant to be used for classification tasks using transfer learning with pre-trained deep neural networks. The method we introduce comes with the following interests:
\begin{itemize}
    \item It is able to incrementally learn new classes, while achieving competing accuracy with off-the-shelf non-incremental transfer methods
    \item It performs learning with a very limited complexity compared to existing counterparts, making it a competitive solution for embedded devices
    \item It builds on top of well-known models of Associative Memories and Self-Organizing Maps, each with its own set of hyperparameters that can be advantageously tuned in order to adapt to the ad-hoc constraints of a given problem
\end{itemize}
We experimentally prove these points in Section~\ref{Experiments} by performing experiments using competitive vision transfer benchmarks.

\section{Self-Organizing Maps and Sparse Associative Memories}

\subsection{Self-Organizing Maps}

\subsubsection{Presentation} 
\begin{figure}
  \centering
  \resizebox{0.53\columnwidth}{!}{%
    \begin{tikzpicture}
      
      \tikzstyle{every node} = [draw, circle, minimum width=2pt, inner sep = 2pt, fill=white]

      \foreach \x/\delta in {0/0,0.3/0.5,0.6/1,0.9/1.5,1.2/2}{
        \draw[opacity=0.2]
        (\x,\delta) -- (\x,\delta+2);
      }

      \foreach \y in {0,0.5,1,1.5,2}{         
        \draw[opacity=0.2]
        (0,\y) -- (1.2,2+\y);
      }

      \foreach \x/\delta in {0/0,0.3/0.5,0.6/1,0.9/1.5,1.2/2}{
        \foreach \y in {0,0.5,1,1.5,2}{
          \node[opacity=0.2] at (\x,\y+\delta) {};
        }
      }
      \node (center) at (0.6,2) {};
      \draw[opacity=0.2]
      (-0.2,-0.8) -- (1.5,2) -- (1.5,4.9) -- (-0.2, 2.1) -- cycle;

      \foreach \y in {0,...,10}{
        \node(\y) at (-2,.5*\y-1) {};
        \path[opacity=0.5,dashed]
        (center) edge (\y);
      }

      \tikzstyle{every node} = []
      \node at (-3.2,2.5) {\small{Input neuron $j$}};
      \node[fill opacity=0.9, fill=white] at (2,1.8) {\small{Output neuron $i$}};
      \node[fill=white] at (-1,2.25) {\tiny{$\mathbf{w}_{i}(j)$}};
      
    \end{tikzpicture}}
  
  \caption{Depiction of a self-organizing map layer.}
  \label{SOM}
\end{figure}
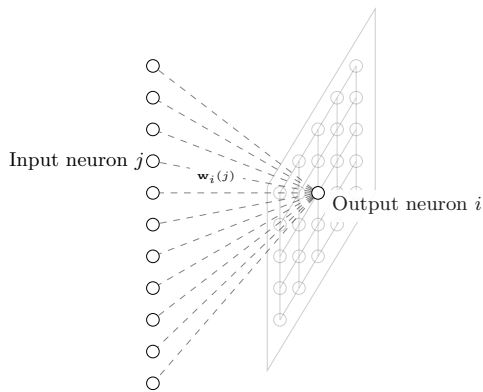

A Self-Organizing Map (SOM)~\cite{58325} is a fully connected layer of $N$ neurons that associates a $d$-dimensional input vector $\mathbf{x}$ with an $N$-dimensional output vector $\mathbf{q(x)}$. These neurons are organized on a 2D-grid of $q$ by $r$ units in a way that each neuron but those on the edges has 4 direct neighbors. All these neurons are entirely connected with the $d$ neurons of the previous layer. The weights corresponding to a given neuron $i$ are denoted $\mathbf{w}_{i}$. Figure~\ref{SOM} depicts an example of such an input layer and a map layer.

\subsubsection{Inference} When an input vector $\mathbf{x}$ is presented to the map layer, the corresponding output is computed as a vector $q(x)\in\{0,1\}^{N}$ containing a single 1. The coordinate $i^*$ which value is 1 is defined as $i^*=\arg\min_{i=1}^{N}{dist(\mathbf{w}_{i}, x)}$. This $i^*$-th neuron is referred to as the Best-Matching Unit (BMU). 
Note that when the vector $x$ and the vectors $\mathbf{w}_{i}$ all have a unit norm, the dynamics of self-organizing maps is equivalent to:

\begin{equation}
  \mathbf{q(x)} = h(W \cdot \mathbf{x})\;,
  \label{eq:feedforward}
\end{equation}
where $h$ is a Winner-Takes-All (WTA) operator (all values are put to 0 except for the maximal one, which is put to 1) and $W$ is the matrix which lines are the vectors $\mathbf{w}_{i}$.

\subsubsection{Training} In contrast with classic fully-connected layers used in deep neural networks, SOMs are built so that neighbor units contain strong inner dependencies, as explained below. The learning algorithm is performed for a specific number of epoch E and a specific batch size. The parameters $\mathbf{w}_{i}$ are first initialized at randoms. The learning procedure is then performed by iterating the following operations for t from 0 to E:
\begin{enumerate}
\item The training set $\mathcal{X}$ is randomly shuffled.
\item For each input vector $\mathbf{x}$ in $\mathcal{X}$, the corresponding output vector $\mathbf{y}$ is computed. Denoting $i^*$ as the neuron where $\mathbf{y}(i^*)=1$, we perform the following update of all weights:
\begin{equation}
  \forall i, \mathbf{w}_{i} \leftarrow \mathbf{w}_{i} + (\mathbf{x} - \mathbf{w}_{i}) A(t) \Theta(t, i^*, i)\;.
\end{equation}
where $A$ is the learning decay function expressed as $A(t) = \alpha T(t)$ with $\alpha$ is the learning rate and $T$ is a function which decreases with $t$ that can be expressed as $T(t) = 1 - \frac{t}{E}$ or $T(t) = e^{ - \frac{t}{E}}$ ;  $\Theta$ the neighborhood function which decreases with $t$ and the distance in the grid between neurons $i$ and $i'$ defined as $\Theta(t, i, i') = e^{ - \frac{d_{i,i'}^2}{2 (\theta  T(t))^2}}$ where $T$ is the decreasing function defined above and $d_{i,i'}$ is the distance between the $i$-th and the ${i'}$-th neurons in the grid independently of their associated vectors $\mathbf{w}_{i}$ and $\mathbf{w}_{j}$. Note that the latter depends only on the topology of the chosen self-organizing map.
\end{enumerate}

\subsubsection{Quantizing with multiple SOMs} \label{quantSom} A popular way to quantize a vector is to use Product Quantization~\cite{jegou:inria-00514462} (PQ). PQ consists in the following: a) splitting the input vectors into $k$ distinct subparts and b) quantizing each part individually and independently from the others. The term ``product'' comes from the fact that the initial space is divided into a Cartesian product of lower dimensional subspaces.

In this section we propose to use multiple SOMs in order to perform PQ, using one for each subspace. The study is restricted to the case where vectors in each subspace all have the same dimension and where each SOM contains the same number $N$ of neurons, such that the number of anchor vectors in the product (initial) space is $N^k$.

Concretely, let us consider $d k$-dimensional input vectors. Our methodology is summarized as follows:
\begin{itemize}
\item Training:
  \begin{enumerate}
  \item Initialize $k$ SOMs with input dimension $d$ (indexed from 1 to $k$). They each contain $N$ neurons,
  \item Split training vectors regularly into $k$ $d$-dimensional subvectors each. The first subvector of each train vector is used to train the first SOM, the second subvector of each train vector to train the second SOM, etc.
  \end{enumerate}
\item Quantizing:
  \begin{enumerate}
  \item Split the input vector $x$ into the $k$ corresponding subvectors,
  \item For each subvector, obtain the corresponding output subvector using the associated SOM,
  \item Concatenate the output subvectors to obtain a $kN$-dimensional binary vector containing exactly $k$ 1s, denoted $Q(x)$.
  \end{enumerate}
\end{itemize}

\subsection{Sparse Associative Memories}
Sparse Associative Memories~\cite{gripon2011sparse} (SAMs) are neural networks able to store and retrieve sparse patterns from incomplete inputs. They consist of a neural network made of $p$ distinct groups composed of variable numbers of units bound by binary connections. SAMs are able to store patterns which have exactly one active neuron in each group. Although not optimized for this problem, SAMs are also able to retrieve ``close'' patterns, meaning that some initially active neurons are changed during the retrieving procedure to find a stored pattern.

The learning procedure is as follows: The connections between the neurons are initialized empty ; then for each pattern to store, the corresponding connections are added to the network. Since connections are binary (they either exist or not), they are not reinforced if shared by more than a single pattern.

The retrieving procedure starts from a partial pattern, meaning that some of its active neurons are initially not activated. Then, an iterative procedure is started. This procedure consists in finding in each group of neurons, the neuron (or the neurons) that has the maximum number of connections with the active neurons. Hopefully, after a few iterations, the network stabilizes to the stored pattern.

\section{Proposed model: combining SOMs and SAMs}
\subsection{Presentation} 
We propose a new model by combining SAMs with multiple SOMs. The proposed model is a sparse associative memory composed of $p = k+1$ groups, where the $k$ first groups correspond to $k$ self-organizing maps composed of $N$ neurons each, and the last group is the output layer containing $M$ neurons. Since we are only interested in finding the active neuron in the last group, connections between the first $k$ groups are ignored and the retrieving procedure is not iterated.

The proposed model has two hyper-parameters: $k$ the number of SOMs and $N$ the number of neurons on each SOM (as stated before, we assume that each SOM has the same number of neurons).

\subsection{Training}\label{proposedModelTrai} The proposed model is expressed as $k$ weight matrices $W$s corresponding to each SOM and one sparse matrix $\Omega$ representing the connections between the $k$ SOMs and the output layer. We denote $\mathcal{X}$ the training set containing pairs ($\mathbf{x}$, $\mathbf{y}$), where $\mathbf{x}$ is a $dk$-dimensional vector and $\mathbf{y}$ is an integer value between 1 and $M$ corresponding to the label. The learning procedure consists in two distinct steps:
  \begin{enumerate}
  \item Training the multiple SOMs as described in the section~\ref{quantSom} on the training dataset $\mathcal{X}$ in order to compute the weight matrix $W$ associated with each SOM.
  \item The sparse matrix $\Omega$ is made of $M$ lines and $kN$ columns, initially containing only 0s. Two methods are proposed in order to train $\Omega$:
  \begin{itemize} 
\item \textbf{Binary method:} the sparse matrix $\Omega$ is a binary matrix containing only 1s and 0s. Then for each couple ($\mathbf{x}$, $\mathbf{y}$) in the training set $\mathcal{X}$, the following equation is obtained:
\begin{equation}
\Omega = \max_{(x,y)\in\mathcal{X}}{\mathbf{e}_{y} \cdot Q(\mathbf{x})^\top}
\end{equation}
where $^\top$ is the transpose operator, the $\max$ is applied componentwise, $\mathbf{e}_\ell$ is the vector of size $M$ containing only 0s except for one 1 at the $\ell$-th coordinate and $Q$ is the quantization function which uses the $k$ SOMs of the model as defined in the section~\ref{quantSom}. 
\item \textbf{Integer method:} the sparse matrix $\Omega$ is an integer matrix containing only positive integer values. Then for each couple ($\mathbf{x}$, $\mathbf{y}$) in the training set $\mathcal{X}$, the following equation is obtained:
\begin{equation}
\Omega = \sum_{(x,y)\in\mathcal{X}}{\mathbf{e}_{y} \cdot Q(\mathbf{x})^\top}
\end{equation}
where $^\top$ is the transpose operator, $\mathbf{e}_\ell$ is the vector of size $M$ containing only 0s except for one 1 at the $\ell$-th coordinate and $Q$ is the quantization function which uses the $k$ SOMs of the model as defined in the section~\ref{quantSom}. 
\end{itemize}
  \end{enumerate}

For reasons of clarity, we denote $k$x$N$-AL the proposed model composed of $k$ SOMs with $N$ neurons on each SOM if it was trained using the binary method and $k$x$N$-IAL if it was trained using the integer method.

The binary method is the faster one because it takes advantage of the fact that both $\Omega$ and $Q(x)$ are binary variables, implying that the computation of the product of both can be highly optimized during implementation. However, it is important to notice that a lot of information is lost due the binary representation of elements inside the matrix $\Omega$ and the integer method has been designed to mitigate this loss. Although it is no longer possible to take advantage of binary operations to optimize the implementation, integer operations remain faster than floating-point operations.

\subsection{Inference} The retrieving procedure is also twofold and does not depend on the method used for the training. By considering an input vector $\mathbf{x}'$, the following prediction can be obtained by:
\begin{equation}
h\left(\Omega \cdot Q(\mathbf{x}')\right)\;
\end{equation}
where $h$ is an activation function, the Winner-Takes-All (WTA) function is the most commonly used and $Q$ is the quantization function which uses the $k$ SOMs of the model as defined in the section~\ref{quantSom}. 

When used this way, the sparse associative memory basically emulates a majority vote among the multiple self-organizing maps.

\subsection{Proposed model used as classifier}
\subsubsection{Presentation}
The model composed of multiple self-organizing maps and one sparse associative memory proposed in the previous section has been designed to be used as a new neural network classifying layer for a conventional Deep Neural Network. The proposed model is a neural network classifier that relies on Transfer Learning and it has to be combined with a feature extractor. In general, the feature extractor is a Convolutional Neural Network pre-trained on a universal dataset, such as ImageNet for image classification, which classification layer has been removed.

\subsubsection{Training}
We denote $\mathcal{U}$ the universal training dataset used to train the feature extractor. $\mathcal{U}$ contains pairs ($\mathbf{x}$, $\mathbf{y}$), where $\mathbf{x}$ is a training vector and $\mathbf{y}$ is the corresponding label, and $\mathcal{U'}$ the dataset containing pairs ($\mathbf{x'}$, $\mathbf{y}$) where $\mathbf{x'}$ is a k-dimensional vector obtained as the output when $\mathbf{x}$ is passed as the input to the feature extractor.
Similarly, we denote $\mathcal{T}$ the domain specific training set and $\mathcal{T'}$.

As described in section~\ref{proposedModelTrai}, the first step is to train the SOMs of the proposed model on the universal dataset $\mathcal{U'}$. The second step consists in training from scratch the matrix $\Omega$ using the domain specific training set $\mathcal{T'}$.
The first step, training the SOMs, is the most computationally expensive task and has to be done ahead ; while the second step, the training on the domain specific dataset, takes fully advantage of the fast incremental learning algorithm described above and has been designed to be done in real-time on the edge.

\subsubsection{Application}\label{PMincrem}
The strengths of this learning procedure is that it is simple and fast: once the feature extractor and the self-organizing maps have been trained on the universal dataset, each training element of the specific dataset has to be processed only one time in order to be learned and the procedure is limited to two sparse matrix products and non-linear functions if the input of the model is a unit vector. Moreover, the learning of each element is independent, and thus it is possible to learn new elements in parallel and incrementally. This means that contrary to deep neural networks which rely on the gradient decent algorithm, it is not required to restart the training from scratch in order to learn new classes or new elements for a previously learned class. Therefore in the situation where the proposed model has already been trained on several classes and has to learn a new one, it is just required to process the elements of the new class. While for a standard deep neural network trained using gradient decent algorithm, in order to avoid catastrophic forgetting it is required to restart the learning from scratch and to process again the elements of the previously learned classes in addition to the elements of the new class.

\section{Experiments}\label{Experiments}

\subsection{Protocol}
In the following subsections, the impact of the different hyper-parameters of the proposed model is evaluated and the proposed model is compared with other methods on several small domain-specific datasets. 

In order to be able to express the complete layer as matrix products, inputs of the proposed model are normed: this does not significantly impact the accuracy of the model in this case. 

Because the accuracy and the time required for the training may slightly vary, every experiment has been done thirty times. The mean and the standard deviation are reported for each measure. Measures of the training time only consider the time required to train the classifiers and do not take the time required to extract the features into account.

\subsubsection{Competing methods}
In order to evaluate our method, the proposed model is compared with different classifiers: a brute-force K-Nearest Neighbor classifier, a Support Vector Machine and a simple deep neural network classifier (denoted DNN classifier). The C-SVM with a Gaussian kernel has been used and it has been trained using a One-vs-Rest policy. The deep neural network is composed of three densely connected layers and it has been trained using three different optimizers : stochastic gradient descent, stochastic gradient descent combined with Cyclical Learning rates~\cite{smith2017cyclical} and Adam~\cite{kingma2014adam}. All classifiers, just like the proposed model, are trained using transfer learning with the same feature extractor. The feature extractor used in all the experiments below is a VGG-16 model trained on ILSVRC dataset~\cite{ILSVRC15} whose dense classification layer has been removed, thus the feature vector is a 4096-dimensional floating point vector. In order to compare the four solutions, it has been decided to not use augmentation on training data or fine-tuning of previous layers of the pre-trained VGG-16 model: the use of these techniques will improve the performance of every method.

\subsubsection{Datasets}
The proposed model has been compared with the other classifiers on several domain specific dataset: 102 Category Flower Dataset~\cite{Nilsback08} (denoted by Flower102), the Indoor Scene Recognition Dataset~\cite{quattoni2009recognizing} (denoted by Indoor67), the Caltech-UCSD Birds 200 dataset~\cite{WelinderEtal2010} (denoted by CUB200), the Stanford Dogs Dataset~\cite{KhoslaYaoJayadevaprakashFeiFei_FGVC2011} (denoted by Dog120) and the Stanford 40 Actions Dataset~\cite{yao2011human} (denoted by Stanford40). These datasets contain highly similar images divided into a lot of classes containing only few images each:
\begin{itemize}
    \item Flower102 dataset contains images of 102 different flower species : the training set is composed of 2040 images and the test set is composed of 6149 images.
    \item Indoor67 dataset contains images of 67 different indoor places : the training set is composed of 5360 images and the test set is composed of 1340 images.
    \item CUB200 dataset contains images of 200 different bird species : the training set is composed of 3000 elements and the test set is composed of 3033 elements.
    \item Dog120 dataset contains images of 120 different dog species : the training set is composed of 12000 elements and the test set is composed of 8580 elements.
    \item Stanford40 dataset contains images of 40 distinct type of actions performed by humans : the training set is composed of 4000 elements and the test set is composed of 5532 elements.
\end{itemize}

\subsection{Impact of hyper-parameters}
The first series of experiments consist in comparing the accuracy of the proposed model on the classification of the Stanford Dogs Dataset depending on the two hyper-parameters of the proposed model: $k$ the number of self-organizing maps and $N$ the number of neuron in each self-organizing map.

\begin{figure}[ht!]
\centering
\begin{minipage}[t]{0.47\textwidth}
  \centering
\resizebox{\columnwidth}{!}{%
    \begin{tikzpicture}
\begin{axis}[
	title={},
	xlabel={Learned Classes},
	ylabel={Top-5 Accuracy [\%]},
    ylabel near ticks,
	xmin=0, xmax=120,
	ymin=20, ymax=100,
	legend columns=3,
	legend style={at={(0.5,-0.2)},anchor=north},
	ymajorgrids=true,
	xmajorgrids=true,
	grid style=dashed,
	]

\addplot[
    color=red,
    mark=square,
    mark repeat=5
    ]
coordinates {(1,100.0) (2,100.0) (3,100.0) (4,100.0) (5,100.0) (6,100.0) (7,100.0) (8,99.8588) (9,99.8718) (10,99.7824) (11,99.6982) (12,99.7245) (13,99.7449) (14,99.6792) (15,99.6937) (16,99.5585) (17,99.5763) (18,99.5902) (19,99.604) (20,99.5101) (21,99.4752) (22,99.2786) (23,99.2063) (24,99.2447) (25,99.2636) (26,99.1109) (27,99.0745) (28,99.0125) (29,98.8255) (30,98.5703) (31,98.5771) (32,98.5396) (33,98.4707) (34,98.4364) (35,98.4408) (36,98.452) (37,98.3507) (38,98.3487) (39,98.176) (40,98.1407) (41,98.0475) (42,97.84) (43,97.8705) (44,97.8901) (45,97.8827) (46,97.7261) (47,97.7075) (48,97.5065) (49,97.4156) (50,97.4091) (51,97.2907) (52,96.942) (53,96.9209) (54,96.7038) (55,96.7431) (56,96.7128) (57,96.6375) (58,96.6461) (59,96.5862) (60,96.4524) (61,96.4081) (62,96.3911) (63,96.3908) (64,96.389) (65,96.3459) (66,96.3424) (67,96.3456) (68,96.3028) (69,96.2286) (70,96.2271) (71,96.225) (72,96.204) (73,96.2236) (74,96.1845) (75,96.1644) (76,96.128) (77,96.1109) (78,96.0235) (79,95.9838) (80,95.9888) (81,96.0083) (82,95.9911) (83,95.9912) (84,96.0263) (85,95.9773) (86,95.9609) (87,95.9673) (88,95.9632) (89,95.9911) (90,95.9447) (91,95.9468) (92,95.9171) (93,95.891) (94,95.878) (95,95.8695) (96,95.8148) (97,95.8284) (98,95.8151) (99,95.7901) (100,95.7456) (101,95.7333) (102,95.6171) (103,95.5674) (104,95.5928) (105,95.582) (106,95.5161) (107,95.5199) (108,95.5241) (109,95.553) (110,95.5853) (111,95.5771) (112,95.6089) (113,95.5284) (114,95.4954) (115,95.3926) (116,95.3653) (117,95.3599) (118,95.2961) (119,95.312) (120,95.338) 
};

\addplot[
    color=blue,
    mark=o,
    mark repeat=5
    ]
coordinates {(1,100.0) (2,100.0) (3,100.0) (4,100.0) (5,100.0) (6,100.0) (7,100.0) (8,100.0) (9,99.8718) (10,99.7824) (11,99.6982) (12,99.7245) (13,99.5748) (14,99.5188) (15,99.5406) (16,99.4113) (17,99.2938) (18,99.1803) (19,99.0759) (20,99.1427) (21,99.0671) (22,98.8901) (23,98.8889) (24,98.9426) (25,98.9691) (26,98.6898) (27,98.5456) (28,98.5831) (29,98.4899) (30,98.2435) (31,98.3004) (32,98.0784) (33,98.0231) (34,98.0) (35,98.0156) (36,98.0048) (37,97.9132) (38,97.8534) (39,97.856) (40,97.8308) (41,97.6269) (42,97.4606) (43,97.4446) (44,97.3626) (45,97.3127) (46,97.1108) (47,97.0487) (48,96.9351) (49,96.8782) (50,96.8112) (51,96.6805) (52,96.5086) (53,96.4946) (54,96.4485) (55,96.4679) (56,96.418) (57,96.3685) (58,96.2268) (59,96.0644) (60,95.9794) (61,95.898) (62,95.8875) (63,95.81) (64,95.8145) (65,95.7978) (66,95.7395) (67,95.7299) (68,95.6735) (69,95.626) (70,95.6497) (71,95.653) (72,95.5996) (73,95.644) (74,95.5918) (75,95.5955) (76,95.4917) (77,95.4447) (78,95.4351) (79,95.3673) (80,95.361) (81,95.3516) (82,95.2887) (83,95.2438) (84,95.2854) (85,95.2929) (86,95.2657) (87,95.2163) (88,95.2043) (89,95.2306) (90,95.2059) (91,95.2042) (92,95.1343) (93,95.0692) (94,95.0476) (95,94.9117) (96,94.7759) (97,94.8002) (98,94.7215) (99,94.7088) (100,94.6642) (101,94.6595) (102,94.5735) (103,94.5519) (104,94.5518) (105,94.4341) (106,94.4247) (107,94.4711) (108,94.4529) (109,94.4822) (110,94.4973) (111,94.4838) (112,94.5141) (113,94.4288) (114,94.3419) (115,94.2227) (116,94.1677) (117,94.1463) (118,94.0787) (119,94.09) (120,94.1142) 
};

\addplot[
    color=green,
    mark=triangle,
    mark repeat=5
    ]
coordinates {(1,100.0) (2,100.0) (3,100.0) (4,100.0) (5,100.0) (6,100.0) (7,99.8428) (8,99.5763) (9,99.6154) (10,99.6736) (11,99.5976) (12,99.6327) (13,99.5748) (14,99.4387) (15,99.3874) (16,99.1906) (17,99.0819) (18,98.7705) (19,98.7459) (20,98.5915) (21,98.4257) (22,97.7802) (23,97.7778) (24,97.6838) (25,97.7418) (26,97.2859) (27,97.2675) (28,97.1232) (29,96.896) (30,96.3644) (31,96.3636) (32,96.2721) (33,96.2327) (34,96.2545) (35,96.2084) (36,96.1816) (37,95.961) (38,95.8058) (39,95.392) (40,95.4137) (41,94.9234) (42,94.7752) (43,94.5486) (44,94.281) (45,94.0825) (46,93.7667) (47,93.6759) (48,93.3766) (49,93.1423) (50,93.2237) (51,92.9461) (52,92.4392) (53,92.3496) (54,91.9452) (55,91.9037) (56,91.9293) (57,91.8628) (58,91.7917) (59,91.6938) (60,91.3782) (61,91.2221) (62,91.1666) (63,91.1429) (64,91.1572) (65,91.0475) (66,90.9767) (67,91.003) (68,90.9931) (69,90.9215) (70,90.9721) (71,90.9247) (72,90.916) (73,90.989) (74,90.9983) (75,90.9708) (76,90.9653) (77,90.8895) (78,90.8702) (79,90.6817) (80,90.722) (81,90.7379) (82,90.7315) (83,90.3856) (84,90.4698) (85,90.3689) (86,90.3327) (87,90.2531) (88,90.2955) (89,90.4136) (90,90.3489) (91,90.2537) (92,90.1919) (93,90.1233) (94,90.1102) (95,90.0329) (96,89.8338) (97,89.8942) (98,89.8804) (99,89.8645) (100,89.7837) (101,89.7287) (102,89.5506) (103,89.5019) (104,89.5769) (105,89.5622) (106,89.5595) (107,89.6672) (108,89.722) (109,89.8085) (110,89.8574) (111,89.8) (112,89.8401) (113,89.7618) (114,89.6309) (115,89.3378) (116,89.1257) (117,89.1017) (118,89.0557) (119,89.0847) (120,89.0909) 
};

\addplot[
    color=violet,
    mark=diamond,
    mark repeat=5
    ]
coordinates {(1,100.0) (2,100.0) (3,100.0) (4,100.0) (5,100.0) (6,99.8148) (7,97.4843) (8,97.3164) (9,97.4359) (10,97.8237) (11,97.6861) (12,97.5207) (13,97.534) (14,97.1933) (15,97.3201) (16,96.9095) (17,96.7514) (18,96.3115) (19,96.2376) (20,96.387) (21,95.9184) (22,95.394) (23,94.709) (24,94.713) (25,94.7472) (26,93.4956) (27,93.4773) (28,92.1426) (29,91.6527) (30,90.482) (31,90.6719) (32,90.4689) (33,90.5632) (34,90.4727) (35,89.5819) (36,88.3385) (37,87.1424) (38,86.823) (39,84.8) (40,83.5761) (41,81.7963) (42,80.0058) (43,78.7621) (44,77.7068) (45,77.7416) (46,77.1536) (47,76.7852) (48,76.1299) (49,75.4862) (50,75.0872) (51,73.9566) (52,73.2483) (53,72.7617) (54,72.2145) (55,72.4541) (56,72.5232) (57,72.6743) (58,72.308) (59,71.9069) (60,71.7265) (61,71.0946) (62,71.0029) (63,70.9396) (64,71.03) (65,71.1328) (66,71.1214) (67,71.2215) (68,70.7178) (69,70.3538) (70,69.9134) (71,69.9523) (72,70.0472) (73,70.2935) (74,70.4575) (75,70.5818) (76,70.0963) (77,70.1837) (78,69.918) (79,69.4557) (80,69.6547) (81,69.7944) (82,69.8818) (83,69.3732) (84,69.3551) (85,69.4375) (86,69.2104) (87,68.7184) (88,68.9811) (89,69.4343) (90,69.5064) (91,69.6937) (92,69.5625) (93,69.548) (94,69.6361) (95,69.6199) (96,69.1748) (97,69.3743) (98,69.4226) (99,69.4492) (100,69.3796) (101,69.2569) (102,68.5404) (103,68.5742) (104,68.7576) (105,68.7934) (106,68.8363) (107,69.0923) (108,69.2808) (109,69.5767) (110,69.7349) (111,69.8099) (112,69.7048) (113,69.4808) (114,69.4148) (115,69.1955) (116,68.9581) (117,68.6258) (118,68.6207) (119,68.7698) (120,68.9627) 
};

\addplot[
    color=cyan,
    mark=pentagon,
    mark repeat=5
    ]
coordinates {(1,100.0) (2,100.0) (3,100.0) (4,100.0) (5,100.0) (6,99.4444) (7,93.239) (8,93.9266) (9,94.1026) (10,94.7769) (11,94.9698) (12,95.3168) (13,95.1531) (14,94.8677) (15,92.343) (16,88.7417) (17,86.2288) (18,83.265) (19,83.6304) (20,84.2621) (21,81.9242) (22,79.1898) (23,76.8783) (24,77.5932) (25,78.0069) (26,75.854) (27,75.2314) (28,74.6243) (29,73.6997) (30,72.1814) (31,72.4506) (32,71.2144) (33,71.8389) (34,70.9818) (35,69.2417) (36,67.5267) (37,66.5769) (38,66.0172) (39,64.48) (40,62.9067) (41,61.3097) (42,59.8365) (43,58.4327) (44,57.2182) (45,56.0261) (46,55.4575) (47,54.888) (48,54.2857) (49,53.7871) (50,53.438) (51,52.4042) (52,51.6976) (53,51.1132) (54,50.1857) (55,50.4128) (56,50.2834) (57,50.4595) (58,50.3751) (59,49.9674) (60,49.5377) (61,48.9054) (62,48.6991) (63,48.2472) (64,48.3586) (65,47.9294) (66,47.6286) (67,47.1102) (68,46.6667) (69,46.1314) (70,45.794) (71,45.4719) (72,45.5524) (73,45.9338) (74,46.3234) (75,46.3938) (76,45.9553) (77,45.8408) (78,45.6669) (79,45.1119) (80,45.4308) (81,45.3084) (82,45.1773) (83,44.7936) (84,44.8897) (85,44.8005) (86,44.4959) (87,43.9184) (88,43.4846) (89,42.8458) (90,42.5338) (91,41.8626) (92,41.5503) (93,41.2114) (94,41.4465) (95,41.2152) (96,40.9172) (97,41.3484) (98,41.1782) (99,40.7584) (100,40.5094) (101,40.3645) (102,39.8358) (103,39.893) (104,39.8675) (105,39.8025) (106,39.8422) (107,40.2305) (108,40.7677) (109,40.7533) (110,40.8079) (111,40.9368) (112,40.6027) (113,40.3543) (114,40.1894) (115,39.9952) (116,39.7365) (117,39.536) (118,39.6171) (119,39.9131) (120,40.2681) 
};

\addplot[
    color=orange,
    mark=x,
    mark repeat=5
    ]
coordinates {(1,100.0) (2,100.0) (3,100.0) (4,100.0) (5,100.0) (6,97.7778) (7,88.0503) (8,88.8418) (9,89.1026) (10,90.4244) (11,90.7445) (12,90.8173) (13,91.0714) (14,90.6175) (15,90.1991) (16,86.8285) (17,83.3333) (18,80.6011) (19,78.4158) (20,77.2811) (21,75.102) (22,72.0866) (23,70.8995) (24,72.004) (25,71.5268) (26,68.4605) (27,65.8881) (28,64.4053) (29,63.7164) (30,62.1732) (31,60.2372) (32,58.9547) (33,57.404) (34,56.4727) (35,55.0319) (36,53.4228) (37,52.3729) (38,51.7503) (39,50.272) (40,48.683) (41,47.2514) (42,45.9428) (43,44.8041) (44,43.809) (45,43.0239) (46,42.4291) (47,41.9763) (48,41.3247) (49,40.7114) (50,39.7857) (51,38.9553) (52,38.4782) (53,38.0862) (54,37.3259) (55,37.6606) (56,37.2931) (57,37.1217) (58,36.5843) (59,36.16) (60,35.7988) (61,35.3666) (62,34.935) (63,34.5364) (64,34.325) (65,33.9626) (66,33.6013) (67,33.2075) (68,32.8614) (69,32.4844) (70,32.1655) (71,31.878) (72,31.5581) (73,31.8938) (74,31.9133) (75,31.7122) (76,31.4306) (77,31.2387) (78,30.9736) (79,30.5971) (80,30.3802) (81,30.1711) (82,29.964) (83,29.7095) (84,29.5841) (85,29.3106) (86,29.068) (87,28.6694) (88,28.3707) (89,27.8403) (90,27.6171) (91,27.1813) (92,26.9992) (93,26.8148) (94,27.027) (95,26.8782) (96,26.6548) (97,26.4248) (98,26.2613) (99,25.9948) (100,25.683) (101,25.5157) (102,25.1287) (103,24.9074) (104,24.7127) (105,24.5462) (106,24.2341) (107,23.9156) (108,23.7184) (109,23.482) (110,23.3242) (111,23.27) (112,23.0258) (113,22.8833) (114,22.7538) (115,22.6149) (116,22.4671) (117,22.3557) (118,22.4087) (119,22.3241) (120,22.669) 
};

\legend{ 128x100-AL, 64x100-AL, 32x100-AL, 16x100-AL, 8x100-AL, 4x100-AL}

\end{axis}
\end{tikzpicture}
}
  \caption{Comparison of the Top-5 accuracy of the proposed model on classification task on Dog120 depending on the number of self-organizing maps composing the model.}
  \label{quantLayerCompar}
\end{minipage}\qquad
\begin{minipage}[t]{0.47\textwidth}
  \centering
\resizebox{\columnwidth}{!}{%
\begin{tikzpicture}
\begin{axis}[
	title={},
	xlabel={Learned Classes},
	ylabel={Top-5 Accuracy [\%]},
    ylabel near ticks,
	xmin=0, xmax=120,
	ymin=70, ymax=100,
	legend columns=2,
	legend style={minimum width=3cm},
	legend style={at={(0.5,-0.2)},anchor=north},
	ymajorgrids=true,
	xmajorgrids=true,
	grid style=dashed,
	]

\addplot[
    color=red,
    mark=square,
    mark repeat=5
    ]
coordinates {(1,100.0) (2,100.0) (3,100.0) (4,100.0) (5,100.0) (6,100.0) (7,100.0) (8,99.8588) (9,99.8718) (10,99.7824) (11,99.5976) (12,99.5409) (13,99.4898) (14,99.4387) (15,99.3874) (16,99.2642) (17,99.2232) (18,99.1803) (19,99.0759) (20,99.1427) (21,99.0087) (22,98.8346) (23,98.5714) (24,98.4894) (25,98.4782) (26,98.0346) (27,97.8405) (28,97.6385) (29,97.5252) (30,97.0588) (31,97.1146) (32,96.887) (33,96.8668) (34,96.8727) (35,96.8816) (36,96.8008) (37,96.7014) (38,96.6314) (39,96.448) (40,96.4983) (41,95.7345) (42,95.5341) (43,95.5139) (44,95.4192) (45,95.4669) (46,95.1578) (47,95.0988) (48,94.8831) (49,94.8311) (50,94.7185) (51,94.4838) (52,94.197) (53,94.1497) (54,93.9183) (55,93.9679) (56,94.015) (57,93.9924) (58,93.8217) (59,93.6725) (60,93.4853) (61,93.5175) (62,93.4536) (63,93.3624) (64,93.373) (65,93.3212) (66,93.2677) (67,93.287) (68,93.2153) (69,93.1376) (70,93.1665) (71,93.1935) (72,93.1823) (73,93.2137) (74,93.2024) (75,93.1364) (76,92.9831) (77,92.96) (78,92.9565) (79,92.866) (80,92.902) (81,92.8288) (82,92.8388) (83,92.7637) (84,92.7934) (85,92.8059) (86,92.7661) (87,92.6367) (88,92.6691) (89,92.7745) (90,92.7067) (91,92.7908) (92,92.7245) (93,92.6952) (94,92.6619) (95,92.5771) (96,92.5052) (97,92.5676) (98,92.5926) (99,92.6038) (100,92.5726) (101,92.4696) (102,92.3055) (103,92.1778) (104,92.2266) (105,92.1516) (106,92.163) (107,92.231) (108,92.2724) (109,92.328) (110,92.3462) (111,92.322) (112,92.337) (113,92.1808) (114,92.12) (115,91.9913) (116,91.9401) (117,91.9215) (118,91.8922) (119,91.9163) (120,91.9464) 
};

\addplot[
    color=blue,
    mark=o,
    mark repeat=5
    ]
coordinates {(1,100.0) (2,100.0) (3,100.0) (4,100.0) (5,100.0) (6,100.0) (7,99.8428) (8,99.5763) (9,99.6154) (10,99.6736) (11,99.5976) (12,99.6327) (13,99.5748) (14,99.4387) (15,99.3874) (16,99.1906) (17,99.0819) (18,98.7705) (19,98.7459) (20,98.5915) (21,98.4257) (22,97.7802) (23,97.7778) (24,97.6838) (25,97.7418) (26,97.2859) (27,97.2675) (28,97.1232) (29,96.896) (30,96.3644) (31,96.3636) (32,96.2721) (33,96.2327) (34,96.2545) (35,96.2084) (36,96.1816) (37,95.961) (38,95.8058) (39,95.392) (40,95.4137) (41,94.9234) (42,94.7752) (43,94.5486) (44,94.281) (45,94.0825) (46,93.7667) (47,93.6759) (48,93.3766) (49,93.1423) (50,93.2237) (51,92.9461) (52,92.4392) (53,92.3496) (54,91.9452) (55,91.9037) (56,91.9293) (57,91.8628) (58,91.7917) (59,91.6938) (60,91.3782) (61,91.2221) (62,91.1666) (63,91.1429) (64,91.1572) (65,91.0475) (66,90.9767) (67,91.003) (68,90.9931) (69,90.9215) (70,90.9721) (71,90.9247) (72,90.916) (73,90.989) (74,90.9983) (75,90.9708) (76,90.9653) (77,90.8895) (78,90.8702) (79,90.6817) (80,90.722) (81,90.7379) (82,90.7315) (83,90.3856) (84,90.4698) (85,90.3689) (86,90.3327) (87,90.2531) (88,90.2955) (89,90.4136) (90,90.3489) (91,90.2537) (92,90.1919) (93,90.1233) (94,90.1102) (95,90.0329) (96,89.8338) (97,89.8942) (98,89.8804) (99,89.8645) (100,89.7837) (101,89.7287) (102,89.5506) (103,89.5019) (104,89.5769) (105,89.5622) (106,89.5595) (107,89.6672) (108,89.722) (109,89.8085) (110,89.8574) (111,89.8) (112,89.8401) (113,89.7618) (114,89.6309) (115,89.3378) (116,89.1257) (117,89.1017) (118,89.0557) (119,89.0847) (120,89.0909) 
};

\addplot[
    color=green,
    mark=triangle,
    mark repeat=5
    ]
coordinates {(1,100.0) (2,100.0) (3,100.0) (4,100.0) (5,100.0) (6,100.0) (7,99.3711) (8,99.1525) (9,98.8462) (10,98.9119) (11,98.5915) (12,98.2553) (13,97.9592) (14,97.3537) (15,97.1669) (16,96.3944) (17,95.4096) (18,94.8087) (19,94.4554) (20,94.5499) (21,94.4023) (22,93.7847) (23,93.2804) (24,93.4038) (25,93.3726) (26,92.2789) (27,92.2433) (28,91.7561) (29,91.401) (30,90.8497) (31,90.9486) (32,90.7763) (33,90.8616) (34,90.6182) (35,90.2551) (36,90.1617) (37,89.9697) (38,89.9273) (39,89.472) (40,89.4639) (41,88.6452) (42,88.1203) (43,87.8478) (44,87.0627) (45,86.9707) (46,86.5971) (47,86.4295) (48,86.0519) (49,85.7984) (50,85.3762) (51,84.867) (52,84.0356) (53,83.657) (54,83.1708) (55,83.1193) (56,83.0424) (57,82.8962) (58,82.4139) (59,82.0831) (60,81.5094) (61,81.2752) (62,81.1372) (63,80.9376) (64,80.8781) (65,80.6131) (66,80.4662) (67,80.3774) (68,79.8427) (69,79.5101) (70,79.4033) (71,79.4471) (72,79.5656) (73,79.7158) (74,79.7555) (75,79.7211) (76,79.3492) (77,79.2582) (78,79.2974) (79,78.9854) (80,79.0548) (81,79.0047) (82,78.9789) (83,78.495) (84,78.4475) (85,78.4844) (86,78.2321) (87,77.9918) (88,77.9913) (89,78.1493) (90,77.9943) (91,77.4443) (92,77.2678) (93,77.1876) (94,77.1705) (95,77.1925) (96,76.7587) (97,76.7333) (98,76.7862) (99,76.831) (100,76.7501) (101,76.6036) (102,76.2488) (103,76.3963) (104,76.3958) (105,76.2146) (106,76.0158) (107,76.2139) (108,76.3581) (109,76.4802) (110,76.5883) (111,76.5561) (112,76.4822) (113,76.2248) (114,76.117) (115,75.9136) (116,75.7126) (117,75.6454) (118,75.5939) (119,75.6903) (120,75.8275) 
};

\legend{32x256-AL, 32x100-AL, 32x25-AL}

\end{axis}
\end{tikzpicture}
}
  \caption{Comparison of the Top-5 accuracy of the proposed model on classification task on Dog120 depending on the number of neurons on each self-organizing map composing the model.}
  \label{quantNeuronCompar}
\end{minipage}

\begin{minipage}[t]{0.47\textwidth}

  \centering
\resizebox{\columnwidth}{!}{%
    \begin{tikzpicture}
\begin{axis}[
	title={},
	xlabel={Learned Classes},
	ylabel={Top-5 Accuracy [\%]},
    ylabel near ticks,
	xmin=0, xmax=120,
	ymin=90, ymax=100,
	legend columns=2,
	legend style={minimum width=3cm},
	legend style={at={(0.5,-0.2)},anchor=north},
	ymajorgrids=true,
	xmajorgrids=true,
	grid style=dashed,
	]

\addplot[
    color=red,
    mark=square,
    mark repeat=5
    ]
coordinates {(1,100.0) (2,100.0) (3,100.0) (4,100.0) (5,100.0) (6,100.0) (7,100.0) (8,99.8588) (9,99.8718) (10,99.7824) (11,99.6982) (12,99.7245) (13,99.7449) (14,99.6792) (15,99.6937) (16,99.5585) (17,99.5763) (18,99.5902) (19,99.604) (20,99.5101) (21,99.4752) (22,99.2786) (23,99.2063) (24,99.2447) (25,99.2636) (26,99.1109) (27,99.0745) (28,99.0125) (29,98.8255) (30,98.5703) (31,98.5771) (32,98.5396) (33,98.4707) (34,98.4364) (35,98.4408) (36,98.452) (37,98.3507) (38,98.3487) (39,98.176) (40,98.1407) (41,98.0475) (42,97.84) (43,97.8705) (44,97.8901) (45,97.8827) (46,97.7261) (47,97.7075) (48,97.5065) (49,97.4156) (50,97.4091) (51,97.2907) (52,96.942) (53,96.9209) (54,96.7038) (55,96.7431) (56,96.7128) (57,96.6375) (58,96.6461) (59,96.5862) (60,96.4524) (61,96.4081) (62,96.3911) (63,96.3908) (64,96.389) (65,96.3459) (66,96.3424) (67,96.3456) (68,96.3028) (69,96.2286) (70,96.2271) (71,96.225) (72,96.204) (73,96.2236) (74,96.1845) (75,96.1644) (76,96.128) (77,96.1109) (78,96.0235) (79,95.9838) (80,95.9888) (81,96.0083) (82,95.9911) (83,95.9912) (84,96.0263) (85,95.9773) (86,95.9609) (87,95.9673) (88,95.9632) (89,95.9911) (90,95.9447) (91,95.9468) (92,95.9171) (93,95.891) (94,95.878) (95,95.8695) (96,95.8148) (97,95.8284) (98,95.8151) (99,95.7901) (100,95.7456) (101,95.7333) (102,95.6171) (103,95.5674) (104,95.5928) (105,95.582) (106,95.5161) (107,95.5199) (108,95.5241) (109,95.553) (110,95.5853) (111,95.5771) (112,95.6089) (113,95.5284) (114,95.4954) (115,95.3926) (116,95.3653) (117,95.3599) (118,95.2961) (119,95.312) (120,95.338) 
};

\addplot[
    color=blue,
    mark=o,
    mark repeat=5
    ]
coordinates {(1,100.0) (2,100.0) (3,100.0) (4,100.0) (5,100.0) (6,100.0) (7,100.0) (8,100.0) (9,99.8718) (10,99.7824) (11,99.6982) (12,99.7245) (13,99.5748) (14,99.5188) (15,99.5406) (16,99.4113) (17,99.2938) (18,99.1803) (19,99.0759) (20,99.1427) (21,99.0671) (22,98.8901) (23,98.8889) (24,98.9426) (25,98.9691) (26,98.6898) (27,98.5456) (28,98.5831) (29,98.4899) (30,98.2435) (31,98.3004) (32,98.0784) (33,98.0231) (34,98.0) (35,98.0156) (36,98.0048) (37,97.9132) (38,97.8534) (39,97.856) (40,97.8308) (41,97.6269) (42,97.4606) (43,97.4446) (44,97.3626) (45,97.3127) (46,97.1108) (47,97.0487) (48,96.9351) (49,96.8782) (50,96.8112) (51,96.6805) (52,96.5086) (53,96.4946) (54,96.4485) (55,96.4679) (56,96.418) (57,96.3685) (58,96.2268) (59,96.0644) (60,95.9794) (61,95.898) (62,95.8875) (63,95.81) (64,95.8145) (65,95.7978) (66,95.7395) (67,95.7299) (68,95.6735) (69,95.626) (70,95.6497) (71,95.653) (72,95.5996) (73,95.644) (74,95.5918) (75,95.5955) (76,95.4917) (77,95.4447) (78,95.4351) (79,95.3673) (80,95.361) (81,95.3516) (82,95.2887) (83,95.2438) (84,95.2854) (85,95.2929) (86,95.2657) (87,95.2163) (88,95.2043) (89,95.2306) (90,95.2059) (91,95.2042) (92,95.1343) (93,95.0692) (94,95.0476) (95,94.9117) (96,94.7759) (97,94.8002) (98,94.7215) (99,94.7088) (100,94.6642) (101,94.6595) (102,94.5735) (103,94.5519) (104,94.5518) (105,94.4341) (106,94.4247) (107,94.4711) (108,94.4529) (109,94.4822) (110,94.4973) (111,94.4838) (112,94.5141) (113,94.4288) (114,94.3419) (115,94.2227) (116,94.1677) (117,94.1463) (118,94.0787) (119,94.09) (120,94.1142) 
};

\addplot[
    color=green,
    mark=triangle,
    mark repeat=5
    ]
coordinates {(1,100.0) (2,100.0) (3,100.0) (4,100.0) (5,100.0) (6,100.0) (7,100.0) (8,99.8588) (9,99.8718) (10,99.7824) (11,99.5976) (12,99.5409) (13,99.4898) (14,99.4387) (15,99.3874) (16,99.2642) (17,99.2232) (18,99.1803) (19,99.0759) (20,99.1427) (21,99.0087) (22,98.8346) (23,98.5714) (24,98.4894) (25,98.4782) (26,98.0346) (27,97.8405) (28,97.6385) (29,97.5252) (30,97.0588) (31,97.1146) (32,96.887) (33,96.8668) (34,96.8727) (35,96.8816) (36,96.8008) (37,96.7014) (38,96.6314) (39,96.448) (40,96.4983) (41,95.7345) (42,95.5341) (43,95.5139) (44,95.4192) (45,95.4669) (46,95.1578) (47,95.0988) (48,94.8831) (49,94.8311) (50,94.7185) (51,94.4838) (52,94.197) (53,94.1497) (54,93.9183) (55,93.9679) (56,94.015) (57,93.9924) (58,93.8217) (59,93.6725) (60,93.4853) (61,93.5175) (62,93.4536) (63,93.3624) (64,93.373) (65,93.3212) (66,93.2677) (67,93.287) (68,93.2153) (69,93.1376) (70,93.1665) (71,93.1935) (72,93.1823) (73,93.2137) (74,93.2024) (75,93.1364) (76,92.9831) (77,92.96) (78,92.9565) (79,92.866) (80,92.902) (81,92.8288) (82,92.8388) (83,92.7637) (84,92.7934) (85,92.8059) (86,92.7661) (87,92.6367) (88,92.6691) (89,92.7745) (90,92.7067) (91,92.7908) (92,92.7245) (93,92.6952) (94,92.6619) (95,92.5771) (96,92.5052) (97,92.5676) (98,92.5926) (99,92.6038) (100,92.5726) (101,92.4696) (102,92.3055) (103,92.1778) (104,92.2266) (105,92.1516) (106,92.163) (107,92.231) (108,92.2724) (109,92.328) (110,92.3462) (111,92.322) (112,92.337) (113,92.1808) (114,92.12) (115,91.9913) (116,91.9401) (117,91.9215) (118,91.8922) (119,91.9163) (120,91.9464) 
};

\legend{128x100-AL (12800), 64x100-AL (6400), 32x256-AL (8192)}

\end{axis}
\end{tikzpicture}}

  \caption{Comparison of the Top-5 accuracy of the proposed model on classification task on Dog120 depending on the total number of neurons.}
  \label{neuronCompar}
  \end{minipage}

\end{figure}

Figure~\ref{quantLayerCompar} and Figure~\ref{quantNeuronCompar} compare the Top-5 accuracy of different versions of the proposed model on the classification of Dog120. Figure~\ref{quantLayerCompar} compares versions of the proposed model which each has a different number of SOMs and Figure~\ref{quantNeuronCompar} compares versions of the proposed model which each has a different number of neurons in each SOM. It appears that, both the increase of the number of SOMs composing the model and the increase of the number of neurons in each SOM will increase the accuracy of the model. This is tied to the fact that they will reduce the loss of information induced by the quantization. However, the computational cost of both the training procedure and the inference procedure of the proposed model is directly correlated to the total number of neurons in the model: the sum of the number of neurons in each SOM. Thus, increasing the number of SOMs or the number of neurons per SOM will also increase the cost. It should be noted that increasing the number of neurons is not necessarily synonymous with increasing the accuracy of the model. As shown in Figure~\ref{neuronCompar}, the 32x256-AL is outperformed by the 64x100-AL even though the latter has 28\% less neurons than the first one. In general, for a fixed number of neurons, the model using the largest number of SOMs should be preferred.

\begin{figure}[ht!]
\centering
\begin{minipage}[t]{0.47\textwidth}
  \centering
\resizebox{\columnwidth}{!}{%
\begin{tikzpicture}
\begin{axis}[
	title={},
	xlabel={Learned Classes},
	ylabel={Accuracy [\%]},
    ylabel near ticks,
	xmin=0, xmax=1000,
	ymin=0, ymax=100,
	legend columns=2,
	legend style={minimum width=3cm},
	legend style={at={(0.5,-0.2)},anchor=north},
	ymajorgrids=true,
	xmajorgrids=true,
	grid style=dashed,
	]

\addplot[
    color=red,
    mark=square,
    mark repeat=5
    ]
coordinates {(1,100) (10,100.0) (20,99.4) (30,99.1333) (40,98.5) (50,97.96) (60,97.4) (70,96.2) (80,94.5) (90,93.8667) (100,93.24) (110,92.7091) (120,92.3) (130,91.8154) (140,91.5429) (150,91.2533) (160,90.95) (170,90.6824) (180,90.5) (190,90.021) (200,89.46) (210,88.4476) (220,87.8273) (230,87.5217) (240,87.05) (250,86.536) (260,86.0846) (270,85.7926) (280,85.3) (290,85.1862) (300,85.04) (310,84.6645) (320,84.3313) (330,84.2606) (340,83.8176) (350,83.3086) (360,83.2444) (370,83.0) (380,82.6474) (390,82.3333) (400,81.935) (410,81.9415) (420,81.6667) (430,81.186) (440,81.0591) (450,80.5067) (460,80.4174) (470,80.1745) (480,80.0125) (490,79.5265) (500,79.092) (510,78.5451) (520,78.5615) (530,77.9472) (540,77.5704) (550,77.2145) (560,76.95) (570,76.6877) (580,76.569) (590,76.1458) (600,76.1233) (610,75.8918) (620,75.5516) (630,75.3048) (640,74.8719) (650,74.5477) (660,74.3727) (670,74.0209) (680,73.7206) (690,73.7044) (700,73.4743) (710,73.2085) (720,72.7944) (730,72.4712) (740,72.1487) (750,71.936) (760,71.6289) (770,71.3662) (780,70.9615) (790,70.5873) (800,70.23) (810,70.0667) (820,69.9146) (830,69.7831) (840,69.6405) (850,69.36) (860,69.2279) (870,69.0391) (880,68.8432) (890,68.7371) (900,68.5067) (910,68.3494) (920,68.1348) (930,68.0538) (940,67.7872) (950,67.6232) (960,67.4312) (970,67.0515) (980,66.8837) (990,66.6081) (1000,66.34) 
};

\addplot[
    color=blue,
    mark=o,
    mark repeat=5
    ]
coordinates {(1,100) (10,93.0) (20,90.3) (30,87.9333) (40,86.05) (50,84.36) (60,79.5) (70,76.2286) (80,73.025) (90,72.7111) (100,70.58) (110,69.9273) (120,69.5) (130,68.4615) (140,67.0286) (150,65.8667) (160,64.8625) (170,64.2823) (180,63.6222) (190,61.6316) (200,59.24) (210,57.5143) (220,56.7909) (230,56.2261) (240,56.1333) (250,55.368) (260,54.2077) (270,53.9926) (280,53.0429) (290,52.9586) (300,52.7067) (310,52.1097) (320,51.4938) (330,51.6909) (340,51.1235) (350,50.3257) (360,49.8111) (370,48.6649) (380,48.5632) (390,47.7385) (400,47.075) (410,47.0585) (420,46.5095) (430,45.8465) (440,45.4364) (450,44.2489) (460,43.887) (470,43.2085) (480,42.9542) (490,42.2367) (500,41.756) (510,40.8824) (520,40.8731) (530,40.3434) (540,40.0704) (550,39.6618) (560,39.5607) (570,39.2175) (580,39.1621) (590,38.0678) (600,38.3533) (610,37.9967) (620,37.7323) (630,37.3651) (640,36.6281) (650,36.1169) (660,35.9242) (670,35.7045) (680,35.4353) (690,34.858) (700,34.8743) (710,34.5437) (720,34.0306) (730,33.7205) (740,33.2595) (750,33.0107) (760,32.7263) (770,32.6052) (780,32.1744) (790,31.7519) (800,31.09) (810,30.8519) (820,30.7073) (830,30.5301) (840,30.4571) (850,30.3059) (860,30.093) (870,29.9011) (880,29.7932) (890,29.6876) (900,29.4667) (910,29.4615) (920,29.337) (930,29.2774) (940,29.1468) (950,28.9853) (960,28.8167) (970,28.2845) (980,27.9469) (990,27.7879) (1000,27.474) 
};

\addplot[
    color=green,
    mark=triangle,
    mark repeat=5
    ]
coordinates {(1,100) (10,99.8) (20,99.3) (30,99.4) (40,98.9) (50,98.4) (60,97.9333) (70,96.9429) (80,96.0) (90,95.5333) (100,94.8) (110,94.6545) (120,94.4667) (130,93.9692) (140,93.7429) (150,93.6) (160,93.2875) (170,93.2706) (180,93.0778) (190,92.5895) (200,92.12) (210,91.6952) (220,91.2909) (230,91.113) (240,90.9583) (250,90.536) (260,90.0692) (270,89.8) (280,89.4357) (290,89.4138) (300,89.36) (310,89.2) (320,89.075) (330,89.1152) (340,88.8588) (350,88.7429) (360,88.6) (370,88.3676) (380,88.1526) (390,87.9795) (400,87.805) (410,87.8) (420,87.581) (430,87.2558) (440,87.2091) (450,86.8444) (460,86.7478) (470,86.5447) (480,86.4292) (490,86.0939) (500,85.792) (510,85.4745) (520,85.5) (530,85.0906) (540,84.9741) (550,84.7927) (560,84.6464) (570,84.4702) (580,84.369) (590,84.1186) (600,84.09) (610,84.0229) (620,83.9129) (630,83.7841) (640,83.55) (650,83.3231) (660,83.2697) (670,83.0418) (680,82.7941) (690,82.7768) (700,82.7029) (710,82.569) (720,82.375) (730,82.1808) (740,82.0324) (750,81.968) (760,81.8605) (770,81.813) (780,81.6282) (790,81.3924) (800,81.225) (810,81.1086) (820,80.9878) (830,80.9036) (840,80.8357) (850,80.7106) (860,80.6698) (870,80.6) (880,80.4568) (890,80.4337) (900,80.2778) (910,80.1604) (920,80.0935) (930,80.0602) (940,79.8574) (950,79.7516) (960,79.6646) (970,79.4144) (980,79.198) (990,79.0606) (1000,78.868)
};

\addplot[
    color=violet,
    mark=diamond,
    mark repeat=5
    ]
coordinates {(1,100) (10,94.8) (20,92.5) (30,91.8667) (40,90.1) (50,88.68) (60,87.1333) (70,84.3143) (80,82.4) (90,81.3333) (100,79.68) (110,79.2909) (120,78.8333) (130,78.1385) (140,77.3429) (150,76.3067) (160,75.7375) (170,75.4588) (180,75.2778) (190,73.8) (200,73.24) (210,72.7143) (220,72.1455) (230,71.6956) (240,71.8167) (250,71.232) (260,70.3308) (270,70.0148) (280,69.3357) (290,69.269) (300,69.4133) (310,69.1548) (320,68.6125) (330,68.7151) (340,68.2353) (350,67.9771) (360,67.7611) (370,67.4378) (380,67.1474) (390,66.8205) (400,66.495) (410,66.4) (420,66.1524) (430,65.5116) (440,65.4) (450,65.0444) (460,64.9304) (470,64.583) (480,64.3583) (490,63.7796) (500,63.436) (510,63.1216) (520,63.0385) (530,62.5434) (540,62.2778) (550,62.0909) (560,61.7679) (570,61.3649) (580,61.1345) (590,60.8102) (600,60.97) (610,60.7967) (620,60.5194) (630,60.3016) (640,60.0469) (650,59.8062) (660,59.6727) (670,59.4388) (680,59.2882) (690,59.058) (700,58.9857) (710,58.6507) (720,58.375) (730,58.1342) (740,57.9243) (750,57.7867) (760,57.4342) (770,57.2597) (780,56.9333) (790,56.6025) (800,56.2075) (810,56.0568) (820,55.9073) (830,55.8072) (840,55.6381) (850,55.5553) (860,55.2791) (870,55.1885) (880,55.0227) (890,54.9326) (900,54.7356) (910,54.6044) (920,54.4978) (930,54.3893) (940,54.2106) (950,54.0568) (960,53.925) (970,53.635) (980,53.451) (990,53.3212) (1000,53.092)
};

\legend{128x100-AL Top-5, 128x100-AL Top-1, 128x100-IAL Top-5, 128x100-IAL Top-1}

\end{axis}
\end{tikzpicture}
}
  \caption{Comparison of the binary (128x100-AL) and integer (128x100-IAL) version of proposed model on classification task on ILSVRC 2012 dataset.}
  \label{binIntComparILSVRC}
\end{minipage}\qquad
\begin{minipage}[t]{0.47\textwidth}
  \centering
\resizebox{\columnwidth}{!}{%
\begin{tikzpicture}
\begin{axis}[
	title={},
	xlabel={Learned Classes},
	ylabel={Accuracy [\%]},
    ylabel near ticks,
	xmin=0, xmax=120,
	ymin=60, ymax=100,
	legend columns=2,
	legend style={minimum width=3cm},
	legend style={at={(0.5,-0.2)},anchor=north},
	ymajorgrids=true,
	xmajorgrids=true,
	grid style=dashed,
	]

\addplot[
    color=red,
    mark=square,
    mark repeat=5
    ]
coordinates {(1,100.0) (2,100.0) (3,100.0) (4,100.0) (5,100.0) (6,100.0) (7,100.0) (8,99.8588) (9,99.8718) (10,99.7824) (11,99.6982) (12,99.7245) (13,99.7449) (14,99.6792) (15,99.6937) (16,99.5585) (17,99.5763) (18,99.5902) (19,99.604) (20,99.5101) (21,99.4752) (22,99.2786) (23,99.2063) (24,99.2447) (25,99.2636) (26,99.1109) (27,99.0745) (28,99.0125) (29,98.8255) (30,98.5703) (31,98.5771) (32,98.5396) (33,98.4707) (34,98.4364) (35,98.4408) (36,98.452) (37,98.3507) (38,98.3487) (39,98.176) (40,98.1407) (41,98.0475) (42,97.84) (43,97.8705) (44,97.8901) (45,97.8827) (46,97.7261) (47,97.7075) (48,97.5065) (49,97.4156) (50,97.4091) (51,97.2907) (52,96.942) (53,96.9209) (54,96.7038) (55,96.7431) (56,96.7128) (57,96.6375) (58,96.6461) (59,96.5862) (60,96.4524) (61,96.4081) (62,96.3911) (63,96.3908) (64,96.389) (65,96.3459) (66,96.3424) (67,96.3456) (68,96.3028) (69,96.2286) (70,96.2271) (71,96.225) (72,96.204) (73,96.2236) (74,96.1845) (75,96.1644) (76,96.128) (77,96.1109) (78,96.0235) (79,95.9838) (80,95.9888) (81,96.0083) (82,95.9911) (83,95.9912) (84,96.0263) (85,95.9773) (86,95.9609) (87,95.9673) (88,95.9632) (89,95.9911) (90,95.9447) (91,95.9468) (92,95.9171) (93,95.891) (94,95.878) (95,95.8695) (96,95.8148) (97,95.8284) (98,95.8151) (99,95.7901) (100,95.7456) (101,95.7333) (102,95.6171) (103,95.5674) (104,95.5928) (105,95.582) (106,95.5161) (107,95.5199) (108,95.5241) (109,95.553) (110,95.5853) (111,95.5771) (112,95.6089) (113,95.5284) (114,95.4954) (115,95.3926) (116,95.3653) (117,95.3599) (118,95.2961) (119,95.312) (120,95.338) 
};

\addplot[
    color=blue,
    mark=o,
    mark repeat=5
    ]
coordinates {(1,100.0) (2,100.0) (3,100.0) (4,97.0414) (5,91.3717) (6,90.5556) (7,90.8805) (8,89.4068) (9,90.1282) (10,91.6213) (11,91.7505) (12,90.3581) (13,89.3707) (14,88.8532) (15,89.0505) (16,87.2701) (17,85.452) (18,83.7432) (19,84.1584) (20,84.507) (21,84.3732) (22,82.3529) (23,82.4868) (24,83.1319) (25,83.3088) (26,82.6392) (27,81.6659) (28,81.9236) (29,81.6695) (30,80.5147) (31,81.0277) (32,81.2068) (33,81.4994) (34,81.6364) (35,81.467) (36,81.2866) (37,81.1511) (38,81.1096) (39,80.416) (40,80.5392) (41,80.2043) (42,79.9766) (43,79.3299) (44,79.4281) (45,79.5331) (46,79.267) (47,78.9987) (48,78.1558) (49,78.0706) (50,77.7529) (51,77.0076) (52,76.6193) (53,76.5277) (54,75.1161) (55,75.2982) (56,75.4477) (57,75.4315) (58,75.1324) (59,74.7771) (60,74.6506) (61,74.5165) (62,74.5699) (63,74.5281) (64,74.5999) (65,74.5432) (66,74.6182) (67,74.717) (68,74.5723) (69,74.4362) (70,74.4947) (71,74.5281) (72,74.5231) (73,74.7242) (74,74.699) (75,74.7477) (76,74.5864) (77,74.4869) (78,74.6434) (79,74.6345) (80,74.7122) (81,74.3563) (82,74.1134) (83,73.8746) (84,73.8508) (85,73.6605) (86,73.5474) (87,73.2408) (88,73.4539) (89,73.7284) (90,73.2788) (91,72.7413) (92,72.6017) (93,72.5765) (94,72.671) (95,72.5681) (96,72.3657) (97,72.5617) (98,72.5868) (99,72.3472) (100,71.6705) (101,71.6445) (102,71.6572) (103,71.6207) (104,71.8535) (105,71.8099) (106,71.6634) (107,71.8633) (108,71.9969) (109,72.1592) (110,72.3112) (111,72.3568) (112,72.4354) (113,72.193) (114,72.1345) (115,71.8249) (116,71.6527) (117,71.7311) (118,71.7173) (119,71.7777) (120,71.9231) 
};

\addplot[
    color=green,
    mark=triangle,
    mark repeat=5
    ]
coordinates {(1,100.0) (2,100.0) (3,100.0) (4,100.0) (5,100.0) (6,100.0) (7,100.0) (8,100.0) (9,99.8718) (10,99.7824) (11,99.5976) (12,99.449) (13,99.4898) (14,99.5188) (15,99.5406) (16,99.5585) (17,99.5763) (18,99.4536) (19,99.4719) (20,99.5101) (21,99.4169) (22,99.1121) (23,98.836) (24,98.7412) (25,98.7727) (26,98.5962) (27,98.5456) (28,98.4543) (29,98.3641) (30,97.9575) (31,98.0237) (32,98.0784) (33,98.0604) (34,97.9636) (35,97.9093) (36,97.9016) (37,97.8122) (38,97.8534) (39,97.792) (40,97.8308) (41,97.717) (42,97.4606) (43,97.473) (44,97.4181) (45,97.367) (46,97.2178) (47,97.1542) (48,96.8831) (49,96.8014) (50,96.7115) (51,96.6317) (52,96.3882) (53,96.3761) (54,96.2628) (55,96.2385) (56,96.2367) (57,96.234) (58,96.0503) (59,95.9774) (60,95.8719) (61,95.8555) (62,95.8246) (63,95.8515) (64,95.835) (65,95.8181) (66,95.7998) (67,95.8093) (68,95.7129) (69,95.6065) (70,95.5919) (71,95.5577) (72,95.543) (73,95.5319) (74,95.5177) (75,95.4671) (76,95.3463) (77,95.2647) (78,95.2746) (79,95.2087) (80,95.204) (81,95.248) (82,95.2373) (83,95.0909) (84,95.1002) (85,95.076) (86,95.0505) (87,95.0694) (88,95.0105) (89,95.0721) (90,95.0016) (91,95.034) (92,94.9962) (93,94.9931) (94,94.8664) (95,94.8369) (96,94.7165) (97,94.7562) (98,94.7652) (99,94.752) (100,94.6642) (101,94.6171) (102,94.4901) (103,94.4147) (104,94.3761) (105,94.2605) (106,94.2012) (107,94.1862) (108,94.1469) (109,94.1925) (110,94.2096) (111,94.1732) (112,94.1943) (113,94.0867) (114,94.0384) (115,93.9332) (116,93.9401) (117,93.9441) (118,93.8896) (119,93.8785) (120,93.9161) 
};

\addplot[
    color=violet,
    mark=diamond,
    mark repeat=5
    ]
coordinates {(1,100.0) (2,97.8102) (3,97.9239) (4,94.6746) (5,88.2743) (6,88.3333) (7,87.8931) (8,87.1469) (9,87.6923) (10,89.445) (11,89.7384) (12,87.8788) (13,87.6701) (14,87.5702) (15,87.4426) (16,85.6512) (17,84.0395) (18,82.7869) (19,83.1683) (20,83.5273) (21,83.4402) (22,81.1321) (23,81.4286) (24,82.0745) (25,82.0815) (26,81.2354) (27,80.476) (28,80.6784) (29,80.4111) (30,79.3301) (31,79.7628) (32,79.7848) (33,80.0448) (34,80.1818) (35,80.085) (36,79.8762) (37,79.8384) (38,79.7886) (39,78.656) (40,78.9278) (41,78.5521) (42,78.2837) (43,77.7115) (44,77.8734) (45,78.013) (46,77.9026) (47,77.6285) (48,76.961) (49,76.74) (50,76.1086) (51,75.3234) (52,74.9097) (53,74.9882) (54,74.1876) (55,74.3119) (56,74.5409) (57,74.5573) (58,73.9409) (59,73.6464) (60,73.5541) (61,73.3688) (62,73.3529) (63,73.3043) (64,73.2868) (65,73.3049) (66,73.2315) (67,73.287) (68,73.1367) (69,72.6672) (70,72.8393) (71,72.9457) (72,72.9745) (73,73.1352) (74,72.995) (75,73.0409) (76,72.8958) (77,72.7944) (78,72.9137) (79,72.9258) (80,73.0555) (81,72.5073) (82,72.4516) (83,72.176) (84,72.0492) (85,71.9246) (86,71.6769) (87,71.3306) (88,71.5324) (89,71.9537) (90,71.6441) (91,70.9004) (92,70.7751) (93,70.7807) (94,70.8138) (95,70.8022) (96,70.4512) (97,70.6669) (98,70.7495) (99,70.6603) (100,69.9915) (101,69.8644) (102,69.7092) (103,69.5622) (104,69.8121) (105,69.5142) (106,69.2965) (107,69.4808) (108,69.7016) (109,69.9295) (110,70.0475) (111,70.1329) (112,70.2829) (113,70.1161) (114,70.0097) (115,69.5694) (116,69.2695) (117,69.3754) (118,69.318) (119,69.4043) (120,69.5455) 
};

\legend{128x100-AL Top-5, 128x100-AL Top-1, 128x100-IAL Top-5, 128x100-IAL Top-1}

\end{axis}
\end{tikzpicture}
}
  \caption{Comparison of the binary (128x100-AL) and integer (128x100-IAL) version of proposed model on classification task on Dog120.}
  \label{binIntComparDOG}
\end{minipage}
\end{figure}

Figure~\ref{binIntComparILSVRC} and Figure~\ref{binIntComparDOG} compare the original proposed model with the improved version which uses integer values instead of binary values on two different classification tasks. The binary model is more accurate than the integer one on the Dog120 dataset while the integer model is more accurate on the ILSVRC dataset. In fact the binary model is more efficient for datasets containing few images per classes whereas the integer model is more efficient on larger datasets containing more images per classes.

\subsection{Comparison with other methods}

\begin{figure}
  \centering
  \begin{minipage}[t]{0.47\textwidth}
  \centering
\resizebox{\columnwidth}{!}{%
    \begin{tikzpicture}
\begin{axis}[
	title={},
	xlabel={Learned Classes},
	ylabel={Accuracy [\%]},
    ylabel near ticks,
	xmin=0, xmax=120,
	ymin=70, ymax=100,
	ymajorgrids=true,
	xmajorgrids=true,
	grid style=dashed,
	legend columns=2,
	legend style={at={(0.5,-0.2)},anchor=north}
	]

\addplot[
    color=red,
    mark=square,
    mark repeat=5
    ]
coordinates {(1,100.0) (2,100.0) (3,100.0) (4,100.0) (5,100.0) (6,100.0) (7,100.0) (8,99.8588) (9,99.8718) (10,99.7824) (11,99.6982) (12,99.7245) (13,99.7449) (14,99.6792) (15,99.6937) (16,99.5585) (17,99.5763) (18,99.5902) (19,99.604) (20,99.5101) (21,99.4752) (22,99.2786) (23,99.2063) (24,99.2447) (25,99.2636) (26,99.1109) (27,99.0745) (28,99.0125) (29,98.8255) (30,98.5703) (31,98.5771) (32,98.5396) (33,98.4707) (34,98.4364) (35,98.4408) (36,98.452) (37,98.3507) (38,98.3487) (39,98.176) (40,98.1407) (41,98.0475) (42,97.84) (43,97.8705) (44,97.8901) (45,97.8827) (46,97.7261) (47,97.7075) (48,97.5065) (49,97.4156) (50,97.4091) (51,97.2907) (52,96.942) (53,96.9209) (54,96.7038) (55,96.7431) (56,96.7128) (57,96.6375) (58,96.6461) (59,96.5862) (60,96.4524) (61,96.4081) (62,96.3911) (63,96.3908) (64,96.389) (65,96.3459) (66,96.3424) (67,96.3456) (68,96.3028) (69,96.2286) (70,96.2271) (71,96.225) (72,96.204) (73,96.2236) (74,96.1845) (75,96.1644) (76,96.128) (77,96.1109) (78,96.0235) (79,95.9838) (80,95.9888) (81,96.0083) (82,95.9911) (83,95.9912) (84,96.0263) (85,95.9773) (86,95.9609) (87,95.9673) (88,95.9632) (89,95.9911) (90,95.9447) (91,95.9468) (92,95.9171) (93,95.891) (94,95.878) (95,95.8695) (96,95.8148) (97,95.8284) (98,95.8151) (99,95.7901) (100,95.7456) (101,95.7333) (102,95.6171) (103,95.5674) (104,95.5928) (105,95.582) (106,95.5161) (107,95.5199) (108,95.5241) (109,95.553) (110,95.5853) (111,95.5771) (112,95.6089) (113,95.5284) (114,95.4954) (115,95.3926) (116,95.3653) (117,95.3599) (118,95.2961) (119,95.312) (120,95.338) 
};

\addplot[
    color=blue,
    mark=o,
    mark repeat=5
    ]
coordinates {(1,100.0) (2,100.0) (3,100.0) (4,97.0414) (5,91.3717) (6,90.5556) (7,90.8805) (8,89.4068) (9,90.1282) (10,91.6213) (11,91.7505) (12,90.3581) (13,89.3707) (14,88.8532) (15,89.0505) (16,87.2701) (17,85.452) (18,83.7432) (19,84.1584) (20,84.507) (21,84.3732) (22,82.3529) (23,82.4868) (24,83.1319) (25,83.3088) (26,82.6392) (27,81.6659) (28,81.9236) (29,81.6695) (30,80.5147) (31,81.0277) (32,81.2068) (33,81.4994) (34,81.6364) (35,81.467) (36,81.2866) (37,81.1511) (38,81.1096) (39,80.416) (40,80.5392) (41,80.2043) (42,79.9766) (43,79.3299) (44,79.4281) (45,79.5331) (46,79.267) (47,78.9987) (48,78.1558) (49,78.0706) (50,77.7529) (51,77.0076) (52,76.6193) (53,76.5277) (54,75.1161) (55,75.2982) (56,75.4477) (57,75.4315) (58,75.1324) (59,74.7771) (60,74.6506) (61,74.5165) (62,74.5699) (63,74.5281) (64,74.5999) (65,74.5432) (66,74.6182) (67,74.717) (68,74.5723) (69,74.4362) (70,74.4947) (71,74.5281) (72,74.5231) (73,74.7242) (74,74.699) (75,74.7477) (76,74.5864) (77,74.4869) (78,74.6434) (79,74.6345) (80,74.7122) (81,74.3563) (82,74.1134) (83,73.8746) (84,73.8508) (85,73.6605) (86,73.5474) (87,73.2408) (88,73.4539) (89,73.7284) (90,73.2788) (91,72.7413) (92,72.6017) (93,72.5765) (94,72.671) (95,72.5681) (96,72.3657) (97,72.5617) (98,72.5868) (99,72.3472) (100,71.6705) (101,71.6445) (102,71.6572) (103,71.6207) (104,71.8535) (105,71.8099) (106,71.6634) (107,71.8633) (108,71.9969) (109,72.1592) (110,72.3112) (111,72.3568) (112,72.4354) (113,72.193) (114,72.1345) (115,71.8249) (116,71.6527) (117,71.7311) (118,71.7173) (119,71.7777) (120,71.9231) 
};

\addplot[
    color=green,
    mark=triangle,
    mark repeat=5
    ]
coordinates {(1,100.0) (2,100.0) (3,100.0) (4,100.0) (5,100.0) (6,99.9703703704) (7,99.8867924528) (8,99.7683615819) (9,99.7435897436) (10,99.6343852013) (11,99.5995975855) (12,99.5133149678) (13,99.5221088435) (14,99.4113873296) (15,99.4333843798) (16,99.4025018396) (17,99.2330508475) (18,99.162568306) (19,99.1260726073) (20,98.9785670545) (21,98.8606413994) (22,98.6748057714) (23,98.6603174603) (24,98.6576032226) (25,98.6067746687) (26,98.5100608329) (27,98.4839136183) (28,98.3623872907) (29,98.2852348993) (30,98.1037581699) (31,98.1043478261) (32,98.0845503459) (33,98.0082058933) (34,98.0334545454) (35,98.0375620128) (36,97.9332645339) (37,97.8532480646) (38,97.8520475561) (39,97.84064) (40,97.8611713666) (41,97.8005407029) (42,97.6141272621) (43,97.6127200454) (44,97.5374791782) (45,97.4799131379) (46,97.3509898341) (47,97.3006587616) (48,97.2649350649) (49,97.1699078813) (50,97.096163428) (51,96.9836465707) (52,96.7893089333) (53,96.6584557082) (54,96.6940575673) (55,96.6697247706) (56,96.5654046701) (57,96.5357543152) (58,96.4686672551) (59,96.5131550337) (60,96.3397118899) (61,96.3162592986) (62,96.2615400755) (63,96.1973656918) (64,96.3084735331) (65,96.2332521315) (66,96.1133440514) (67,96.1966236346) (68,96.249754179) (69,95.9826982893) (70,96.1487006737) (71,96.1067683508) (72,96.0651558074) (73,96.0908581043) (74,96.1321541026) (75,96.0506514957) (76,95.9807307762) (77,95.9079042132) (78,95.9156562054) (79,95.8754623921) (80,95.9020753401) (81,95.9309659582) (82,95.8840157615) (83,95.8646169526) (84,95.8166357973) (85,95.7703221499) (86,95.8301605695) (87,95.8130612245) (88,95.8259325044) (89,95.8762478213) (90,95.7760924238) (91,95.8709777227) (92,95.8177283192) (93,95.7506467813) (94,95.7602295032) (95,95.6311733014) (96,95.6500445236) (97,95.6778789659) (98,95.643773695) (99,95.6751009227) (100,95.5744877632) (101,95.5040265612) (102,95.4755113399) (103,95.3556561914) (104,95.3647875265) (105,95.3572699768) (106,95.2373438527) (107,95.3614743839) (108,95.2843662331) (109,95.3838078441) (110,95.3976988494) (111,95.373752433) (112,95.405904059) (113,95.366320505) (114,95.2570017808) (115,95.3057532264) (116,95.1612774451) (117,95.00297442) (118,95.0659890478) (119,95.0836172796) (120,95.067987568) 
};

\addplot[
    color=violet,
    mark=diamond,
    mark repeat=5
    ]
coordinates {(1,100.0) (2,99.8102189781) (3,99.4532871972) (4,96.0355029585) (5,90.6238938053) (6,90.6111111111) (7,90.4119496856) (8,89.8926553672) (9,90.1615384615) (10,91.3993471164) (11,91.847082495) (12,90.7897153352) (13,89.8588435374) (14,89.4290296712) (15,89.2894333844) (16,87.7954378219) (17,86.040960452) (18,84.9658469945) (19,85.0059405941) (20,85.6472749541) (21,85.1183673469) (22,83.8346281909) (23,83.8433862434) (24,84.390735146) (25,84.2965144821) (26,83.528310716) (27,82.6963420009) (28,83.1215113783) (29,82.8498322148) (30,81.4926470588) (31,81.9280632411) (32,82.2690238278) (33,82.429690414) (34,82.7701818182) (35,82.4670446492) (36,82.1293429652) (37,81.9831706496) (38,82.1195508587) (39,81.74016) (40,81.8450573288) (41,81.4286572544) (42,80.6094570928) (43,80.5241340148) (44,80.5263742365) (45,80.5217155266) (46,80.3488496522) (47,80.0790513834) (48,79.44) (49,79.3403275333) (50,79.1619332337) (51,78.447644618) (52,77.9711052252) (53,77.5945049739) (54,77.3254410399) (55,77.2926605505) (56,77.1698027658) (57,77.2194575207) (58,76.9774933804) (59,77.0028267015) (60,76.9774242098) (61,76.8552603613) (62,76.7897608057) (63,76.7771209293) (64,77.0470865819) (65,77.023954527) (66,76.8659565916) (67,76.9424031777) (68,76.8598820059) (69,76.2402799378) (70,76.8946102021) (71,76.9456625358) (72,76.8385269122) (73,76.9419517667) (74,77.0911279867) (75,77.064599009) (76,76.8982912198) (77,77.0485235866) (78,76.8986269615) (79,76.9032939933) (80,77.1961109173) (81,76.8960601348) (82,76.345725544) (83,76.3457618482) (84,76.1075096818) (85,76.001919546) (86,76.0677040225) (87,75.7959183673) (88,75.9837720006) (89,75.9986531453) (90,75.8770826784) (91,75.5109065594) (92,75.5433614735) (93,75.5531882514) (94,75.4639136343) (95,75.2476803352) (96,75.3491392104) (97,75.4788484136) (98,75.4684310295) (99,75.2905132641) (100,74.5493027888) (101,74.3274936423) (102,74.5944065674) (103,74.5825900005) (104,74.7654454509) (105,74.6316070475) (106,74.2612316459) (107,74.646726229) (108,74.8329507779) (109,74.961787184) (110,75.1138069035) (111,74.9463701495) (112,75.2332923329) (113,75.1097536143) (114,75.0947061681) (115,74.6809793752) (116,74.3608782435) (117,74.2375570097) (118,74.4541622345) (119,74.4699016958) (120,74.6810411811) 
};

\legend{128x100-AL Top-5, 128x100-AL Top-1, DNN classifier Top-5, DNN classifier Top-1}

\end{axis}
\end{tikzpicture}}

  \caption{Comparison of 128x100-AL with the DNN classifier trained using SGD on classification task on the Dog120 dataset depending on the number of classes learned. Both classifiers have been trained on all the considered classes at once.}
  \label{NNIALCompFS}

\end{minipage}\qquad
\begin{minipage}[t]{0.47\textwidth}
  \centering
\resizebox{\columnwidth}{!}{%
  \centering
    \begin{tikzpicture}
\begin{axis}[
	title={},
	xlabel={Learned Classes},
	ylabel={Accuracy [\%]},
    ylabel near ticks,
	xmin=0, xmax=120,
	ymin=0, ymax=100,
	ymajorgrids=true,
	xmajorgrids=true,
	grid style=dashed,
	legend columns=2,
	legend style={at={(0.5,-0.2)},anchor=north}
	]

\addplot[
    color=red,
    mark=square,
    mark repeat=5
    ]
coordinates {(1,100.0) (2,100.0) (3,100.0) (4,100.0) (5,100.0) (6,100.0) (7,100.0) (8,99.8588) (9,99.8718) (10,99.7824) (11,99.6982) (12,99.7245) (13,99.7449) (14,99.6792) (15,99.6937) (16,99.5585) (17,99.5763) (18,99.5902) (19,99.604) (20,99.5101) (21,99.4752) (22,99.2786) (23,99.2063) (24,99.2447) (25,99.2636) (26,99.1109) (27,99.0745) (28,99.0125) (29,98.8255) (30,98.5703) (31,98.5771) (32,98.5396) (33,98.4707) (34,98.4364) (35,98.4408) (36,98.452) (37,98.3507) (38,98.3487) (39,98.176) (40,98.1407) (41,98.0475) (42,97.84) (43,97.8705) (44,97.8901) (45,97.8827) (46,97.7261) (47,97.7075) (48,97.5065) (49,97.4156) (50,97.4091) (51,97.2907) (52,96.942) (53,96.9209) (54,96.7038) (55,96.7431) (56,96.7128) (57,96.6375) (58,96.6461) (59,96.5862) (60,96.4524) (61,96.4081) (62,96.3911) (63,96.3908) (64,96.389) (65,96.3459) (66,96.3424) (67,96.3456) (68,96.3028) (69,96.2286) (70,96.2271) (71,96.225) (72,96.204) (73,96.2236) (74,96.1845) (75,96.1644) (76,96.128) (77,96.1109) (78,96.0235) (79,95.9838) (80,95.9888) (81,96.0083) (82,95.9911) (83,95.9912) (84,96.0263) (85,95.9773) (86,95.9609) (87,95.9673) (88,95.9632) (89,95.9911) (90,95.9447) (91,95.9468) (92,95.9171) (93,95.891) (94,95.878) (95,95.8695) (96,95.8148) (97,95.8284) (98,95.8151) (99,95.7901) (100,95.7456) (101,95.7333) (102,95.6171) (103,95.5674) (104,95.5928) (105,95.582) (106,95.5161) (107,95.5199) (108,95.5241) (109,95.553) (110,95.5853) (111,95.5771) (112,95.6089) (113,95.5284) (114,95.4954) (115,95.3926) (116,95.3653) (117,95.3599) (118,95.2961) (119,95.312) (120,95.338) 
};

\addplot[
    color=blue,
    mark=o,
    mark repeat=5
    ]
coordinates {(1,100.0) (2,100.0) (3,100.0) (4,97.0414) (5,91.3717) (6,90.5556) (7,90.8805) (8,89.4068) (9,90.1282) (10,91.6213) (11,91.7505) (12,90.3581) (13,89.3707) (14,88.8532) (15,89.0505) (16,87.2701) (17,85.452) (18,83.7432) (19,84.1584) (20,84.507) (21,84.3732) (22,82.3529) (23,82.4868) (24,83.1319) (25,83.3088) (26,82.6392) (27,81.6659) (28,81.9236) (29,81.6695) (30,80.5147) (31,81.0277) (32,81.2068) (33,81.4994) (34,81.6364) (35,81.467) (36,81.2866) (37,81.1511) (38,81.1096) (39,80.416) (40,80.5392) (41,80.2043) (42,79.9766) (43,79.3299) (44,79.4281) (45,79.5331) (46,79.267) (47,78.9987) (48,78.1558) (49,78.0706) (50,77.7529) (51,77.0076) (52,76.6193) (53,76.5277) (54,75.1161) (55,75.2982) (56,75.4477) (57,75.4315) (58,75.1324) (59,74.7771) (60,74.6506) (61,74.5165) (62,74.5699) (63,74.5281) (64,74.5999) (65,74.5432) (66,74.6182) (67,74.717) (68,74.5723) (69,74.4362) (70,74.4947) (71,74.5281) (72,74.5231) (73,74.7242) (74,74.699) (75,74.7477) (76,74.5864) (77,74.4869) (78,74.6434) (79,74.6345) (80,74.7122) (81,74.3563) (82,74.1134) (83,73.8746) (84,73.8508) (85,73.6605) (86,73.5474) (87,73.2408) (88,73.4539) (89,73.7284) (90,73.2788) (91,72.7413) (92,72.6017) (93,72.5765) (94,72.671) (95,72.5681) (96,72.3657) (97,72.5617) (98,72.5868) (99,72.3472) (100,71.6705) (101,71.6445) (102,71.6572) (103,71.6207) (104,71.8535) (105,71.8099) (106,71.6634) (107,71.8633) (108,71.9969) (109,72.1592) (110,72.3112) (111,72.3568) (112,72.4354) (113,72.193) (114,72.1345) (115,71.8249) (116,71.6527) (117,71.7311) (118,71.7173) (119,71.7777) (120,71.9231) 
};

\addplot[
    color=green,
    mark=triangle,
    mark repeat=5
    ]
coordinates {(1,100) (2,100.0) (3,100.0) (4,100.0) (5,100.0) (6,86.63522012578616) (7,66.52542372881356) (8,63.333333333333336) (9,56.47442872687704) (10,50.905432595573444) (11,46.372819100091824) (12,44.21768707482993) (13,41.61988773055333) (14,33.61408882082695) (15,30.684326710816777) (16,32.98022598870057) (17,30.259562841530055) (18,30.42904290429043) (19,31.047152480097978) (20,32.59475218658892) (21,28.1354051054384) (22,26.50793650793651) (23,26.283987915407856) (24,23.76043200785469) (25,22.882545624707532) (26,24.548259144997797) (27,22.756547874624303) (28,19.463087248322147) (29,18.66830065359477) (30,18.932806324110672) (31,18.908531898539586) (32,17.5307720999627) (33,14.909090909090908) (34,14.635010630758327) (35,13.48469212246302) (36,11.67956916863009) (37,13.771466314398943) (38,14.624) (39,12.674310505113109) (40,12.916791829378191) (41,11.091652072387625) (42,10.760931289040318) (43,12.354247640199889) (44,11.889250814332248) (45,12.092027822364901) (46,11.989459815546772) (47,11.714285714285714) (48,12.615148413510747) (49,12.506228201295466) (50,11.691481571881864) (51,11.12448832169516) (52,11.653244907626718) (53,10.30640668523677) (54,9.403669724770642) (55,9.113579687145771) (56,9.168347904057386) (57,10.260370697263902) (58,10.023918243096325) (59,9.95484841969469) (60,9.819341126461211) (61,8.665547629039027) (62,8.442231902095001) (63,6.955272876487484) (64,7.206658546488023) (65,7.234726688102894) (66,7.328699106256207) (67,7.944936086529007) (68,8.028771384136858) (69,9.432146294513956) (70,9.323164918970448) (71,8.290840415486308) (72,8.93624976631146) (73,7.59399888868309) (74,7.524316388328134) (75,7.65315397200509) (76,7.598127475693194) (77,7.703281027104137) (78,7.873877047736481) (79,7.830484827345657) (80,8.259892863314326) (81,7.178345040260408) (82,7.117377271955156) (83,7.08873547735309) (84,7.027207477883492) (85,6.969044860122496) (86,7.428571428571429) (87,7.282415630550622) (88,7.875138646807162) (89,7.450487268154668) (90,8.12190594059406) (91,6.983883346124329) (92,7.00045655151423) (93,7.066284161256228) (94,7.108650104759054) (95,6.812110418521817) (96,6.888954171562867) (97,5.832604257801108) (98,6.170703575547866) (99,6.47410358565737) (100,5.8349816332297255) (101,5.523862529567274) (102,5.324550569507342) (103,5.9618764363931325) (104,5.098772023491724) (105,6.548323471400394) (106,6.5130130778195) (107,6.044376434583015) (108,5.681531872008063) (109,5.377688844422211) (110,5.2801590259659585) (111,4.969249692496925) (112,4.593769089798411) (113,4.747450218552696) (114,4.764202146906284) (115,5.269461077844311) (116,5.187388459250446) (117,4.822125044321003) (118,4.723299259781459) (119,4.557109557109557) 
};

\addplot[
    color=violet,
    mark=diamond,
    mark repeat=5
    ]
coordinates {(1,100) (2,17.993079584775085) (3,15.384615384615385) (4,11.504424778761061) (5,9.62962962962963) (6,8.176100628930818) (7,7.344632768361582) (8,6.666666666666667) (9,5.658324265505985) (10,5.231388329979879) (11,4.775022956841139) (12,4.421768707482993) (13,4.170008019246191) (14,3.9816232771822357) (15,3.8263428991905815) (16,3.672316384180791) (17,3.551912568306011) (18,3.432343234323432) (19,3.1843233312921004) (20,3.0320699708454812) (21,2.885682574916759) (22,2.751322751322751) (23,2.618328298086606) (24,2.552773686794305) (25,2.4333177351427233) (26,2.2917584839136182) (27,2.23271790468012) (28,2.1812080536912752) (29,2.1241830065359477) (30,2.0553359683794468) (31,1.9984627209838586) (32,1.939574785527788) (33,1.8909090909090909) (34,1.8426647767540751) (35,1.7887856897144823) (36,1.7502524402558062) (37,1.7173051519154559) (38,1.664) (39,1.6114037806011776) (40,1.5620306398317814) (41,1.5178050204319906) (42,1.4764338444065872) (43,1.4436424208772904) (44,1.4115092290988056) (45,1.3911182450508293) (46,1.370223978919631) (47,1.3506493506493507) (48,1.330603889457523) (49,1.2954658694569008) (50,1.2692213814986575) (51,1.2521069106669878) (52,1.231643770724775) (53,1.2070566388115134) (54,1.1926605504587156) (55,1.1788710043074133) (56,1.165657924232235) (57,1.147396293027361) (58,1.1306805827353772) (59,1.1180391313695979) (60,1.1052072263549415) (61,1.0910616869492236) (62,1.078614395353661) (63,1.0668855149774312) (64,1.0556232237109215) (65,1.045016077170418) (66,1.0327706057596822) (67,1.022615535889872) (68,1.010886469673406) (69,1.0009624639076036) (70,0.9914204003813155) (71,0.9820585457979226) (72,0.9721443260422509) (73,0.9631413224671236) (74,0.9543035419342999) (75,0.9452826758771132) (76,0.9362621534029528) (77,0.927246790299572) (78,0.9159767482825436) (79,0.9068712940355773) (80,0.8985657508208053) (81,0.89086859688196) (82,0.8833021912688975) (83,0.8755682774877925) (84,0.8679686195960608) (85,0.8607846383049164) (86,0.8489795918367347) (87,0.839657678023575) (88,0.8239581682776105) (89,0.817353033637221) (90,0.8044554455445545) (91,0.7981580966999232) (92,0.7913559579972607) (93,0.7851426845840254) (94,0.7782101167315175) (95,0.771742356782428) (96,0.763807285546416) (97,0.758238553514144) (98,0.7497116493656286) (99,0.7398975526465567) (100,0.734670810963549) (101,0.7235285932934465) (102,0.7135995608618086) (103,0.70298769771529) (104,0.6940736785904965) (105,0.6837606837606838) (106,0.6733134792179205) (107,0.6630961489415965) (108,0.655076845553036) (109,0.6503251625812907) (110,0.6460429867064231) (111,0.6396063960639606) (112,0.6353084911423336) (113,0.6313744536182613) (114,0.6271861054155108) (115,0.6227544910179641) (116,0.6186793575252826) (117,0.6145845644722846) (118,0.6109740336035718) (119,0.6060606060606061) 

};

\legend{128x100-AL Top-5, 128x100-AL Top-1, DNN classifier Top-5, DNN classifier Top-1}

\end{axis}
\end{tikzpicture}}

  \caption{Comparison of 128x100-AL with the DNN classifier trained using SGD on classification task on the Dog120 dataset depending on the number of classes learned. Both classifiers have been trained sequentially.}
  \label{NNIALCompINC}

\end{minipage}
\end{figure}

Figure~\ref{NNIALCompFS} compares the accuracy of the proposed model with the accuracy of the deep neural network classifier trained on the Dog120 dataset using stochastic gradient descent depending on the number of learned classes. Both classifier have been trained on all the considered classes at once: this means that each time a new class is added to the training set, the learning procedure is restarted from scratch and the classifier has to learn again the previously learned classes in addition to the new class. It appears that both models have nearly the same accuracy: they have the same Top-5 accuracy and the Top-1 accuracy of proposed model is slightly inferior to the one of the deep neural network classifier. In the case where the number of learned classes equals 120, the loss in accuracy is about 2.7\%: the Top-1 accuracy of the proposed model is 71.9\% and the the Top-1 accuracy of the deep neural network classifier is 74.6\%.

However, as explained in the section~\ref{PMincrem}, contrary to the deep neural network classifier, the proposed model supports incremental learning. Figure~\ref{NNIALCompINC} compares the accuracy of the proposed model with the accuracy of the deep neural network classifier trained on the Dog120 dataset using stochastic gradient descent depending on the number of learned classes. Nevertheless, in this case both classifiers have been trained sequentially: this means that each time a new class is added to the training set, the classifier is only trained on the new elements. As expected, the accuracy of the proposed model is exactly the same because of the learning algorithm used: it does not change if the proposed model is trained on all the dataset at once or sequentially. However, the accuracy of the deep neural network classifier crashed and dropped to 0.66\% for Top-1 accuracy and 4.55\% for Top-5 accuracy in the case where the number of learned classes equals 120 . This is due to catastrophic forgetting: since the deep neural network classifier was trained sequentially, each time it learns a new class it also inevitably forgets all the previously learned classes. Therefore, the deep neural network classifier always predicts the last learned class independently of the input image.

As a result, the proposed model is more flexible than the deep neural network classifier because it supports incremental learning at the cost of the slight decrease of accuracy. The proposed model can be more easily retrained and adapted to a new situation compared to the deep neural network classifier. For example, if the proposed model has already been trained on the first 119 classes of the Dog120 dataset and has to learn the last one, it is just required to process the elements of the new class, while for the deep neural network classifier, it is required to restart the learning from scratch and to process the elements of all the 120 classes of the dataset in order to avoid catastrophic forgetting. In this case, it will be up to approximately 2600 times faster for the proposed model to learn the new class compared to the deep neural network classifier, as shown in Table~\ref{speedCompIncr}. \\

The proposed model has also been compared with the other classifiers on several additional domain specific dataset: 

\begin{table}[ht!]
  \caption{Comparison of the accuracy of 128x100-AL with different transfer learning methods on classification task on different domain specific datasets when learning the complete dataset at once. Values are averaged over 30 runs.}
  \label{transferComp}
  \centering
  \footnotesize
\resizebox{\columnwidth}{!}{%
\def\arraystretch{2}
\begin{tabularx}{\textwidth}{ c c || *{4}{>{\centering\arraybackslash}X}| *{2}{>{\centering\arraybackslash}X} }
\hline
\multicolumn{2}{c ||}{} & \multicolumn{4}{c |}{Non incremental methods} & \multicolumn{2}{c}{Incremental methods} \\
\cline{3-8}
\multicolumn{2}{c ||}{} & DNN classifier (SGD) & DNN classifier (SGD + CycleLR) & DNN classifier (Adam) & SVM & K-NN & Proposed model 128x100-AL  \\
\hline
\multirow{2}{5em}{Flower102} & Top-1 & \textbf{\makecell{75.71 \% \\ \footnotesize{($\pm$ 0.97)}}}   & \makecell{70.52 \% \\ \footnotesize{($\pm$ 2.08)}} & \makecell{72.12 \% \\ \footnotesize{($\pm$ 0.96)}} & \makecell{73.80 \% \\ \footnotesize{($\pm$ 0.0)}} & \makecell{57.13 \% \\ \footnotesize{($\pm$ 0.0)}} & \makecell{66.92 \% \\ \footnotesize{($\pm$ 0.0)}} \\ 
 & Top-5 & \textbf{\makecell{91.89 \% \\ \footnotesize{($\pm$ 0.56)}}}  & \makecell{89.07 \% \\ \footnotesize{($\pm$ 1.25)}} & \makecell{90.17 \% \\ \footnotesize{($\pm$ 0.69)}} & \makecell{90.94 \% \\ \footnotesize{($\pm$ 0.0)}} & \makecell{80.34 \% \\ \footnotesize{($\pm$ 0.0)}} & \makecell{87.48 \% \\ \footnotesize{($\pm$ 0.0)}} \\
\hline
\multirow{2}{5em}{Indoor67} & Top-1 & \makecell{65.43 \% \\ \footnotesize{($\pm$ 0.83)}} & \makecell{64.05 \% \\ \footnotesize{($\pm$ 0.93)}} & \makecell{61.86 \% \\ \footnotesize{($\pm$ 1.48)}} & \textbf{\makecell{67.24 \% \\ \footnotesize{($\pm$ 0.0)}}}  & \makecell{52.76 \% \\ \footnotesize{($\pm$ 0.0)}} & \makecell{56.49 \% \\ \footnotesize{($\pm$ 0.0)}} \\ 
 & Top-5 & \makecell{90.21 \% \\ \footnotesize{($\pm$ 0.50)}} & \makecell{89.57 \% \\ \footnotesize{($\pm$ 0.56)}} & \makecell{87.86 \% \\ \footnotesize{($\pm$ 0.52)}} & \textbf{\makecell{90.90 \% \\ \footnotesize{($\pm$ 0.0)}}} & \makecell{78.36 \% \\ \footnotesize{($\pm$ 0.0)}} & \makecell{87.31 \% \\ \footnotesize{($\pm$ 0.0)}} \\
\hline
\multirow{2}{5em}{CUB200} & Top-1 & \makecell{37.31 \% \\ \footnotesize{($\pm$ 0.97)}} & \makecell{32.69 \% \\ \footnotesize{($\pm$ 1.21)}} & \makecell{31.97 \% \\ \footnotesize{($\pm$ 1.20)}} & \makecell{38.02 \% \\ \footnotesize{($\pm$ 0.0)}} & \makecell{24.70 \% \\ \footnotesize{($\pm$ 0.0)}} & \textbf{\makecell{41.51 \% \\ \footnotesize{($\pm$ 0.0)}}} \\ 
 & Top-5 & \makecell{66.74 \% \\ \footnotesize{($\pm$ 1.05)}} & \makecell{61.83 \% \\ \footnotesize{($\pm$ 1.46)}} & \makecell{61.61 \% \\ \footnotesize{($\pm$ 1.50)}} & \makecell{66.74 \% \\ \footnotesize{($\pm$ 0.0)}} &  \makecell{50.31 \% \\ \footnotesize{($\pm$ 0.0)}} & \textbf{\makecell{76.18 \% \\ \footnotesize{($\pm$ 0.0)}}} \\
\hline
\multirow{2}{5em}{Dog120} & Top-1 & \makecell{74.56 \% \\ \footnotesize{($\pm$ 0.34)}} & \makecell{73.81 \% \\ \footnotesize{($\pm$ 0.29)}} & \makecell{70.80 \% \\ \footnotesize{($\pm$ 0.65)}} & \textbf{\makecell{76.35 \% \\ \footnotesize{($\pm$ 0.0)}}} & \makecell{68.83 \% \\ \footnotesize{($\pm$ 0.0)}} & \makecell{71.91 \% \\ \footnotesize{($\pm$ 0.0)}} \\ 
 & Top-5 & \makecell{95.06 \% \\ \footnotesize{($\pm$ 0.14)}} & \makecell{94.63 \% \\ \footnotesize{($\pm$ 0.15)}} & \makecell{93.37 \% \\ \footnotesize{($\pm$ 0.30)}} & \textbf{\makecell{95.90 \% \\ \footnotesize{($\pm$ 0.0)}}} & \makecell{87.74 \% \\ \footnotesize{($\pm$ 0.0)}} & \makecell{95.34 \% \\ \footnotesize{($\pm$ 0.0)}} \\
 \hline
 \multirow{2}{5em}{Stanford40} & Top-1 & \makecell{68.05 \% \\ \footnotesize{($\pm$ 0.54)}} & \makecell{67.81 \% \\ \footnotesize{($\pm$ 0.42)}} & \makecell{66.21 \% \\ \footnotesize{($\pm$ 0.70)}} & \textbf{\makecell{69.54 \% \\ \footnotesize{($\pm$ 0.0)}}} & \makecell{57.36 \% \\ \footnotesize{($\pm$ 0.0)}} & \makecell{59.06 \% \\ \footnotesize{($\pm$ 0.0)}} \\ 
 & Top-5 & \makecell{90.67 \% \\ \footnotesize{($\pm$ 0.34)}} & \makecell{90.61 \% \\ \footnotesize{($\pm$ 0.28)}} & \makecell{89.46 \% \\ \footnotesize{($\pm$ 0.39)}} & \textbf{\makecell{91.68 \% \\ \footnotesize{($\pm$ 0.0)}}} & \makecell{79.10 \% \\ \footnotesize{($\pm$ 0.0)}} & \makecell{87.02 \% \\ \footnotesize{($\pm$ 0.0)}} \\
\hline
\end{tabularx}}
\end{table}

\begin{table}[ht!]
  \caption{Comparison of the training time of 128x100-AL with different classifiers on different domain specific datasets when learning the complete dataset at once. Training time duration is expressed relatively to the one of the proposed model: the lower is the value, the better it is. Values are averaged over 30 runs.}
  \label{speedComp}
  \centering
\resizebox{\columnwidth}{!}{%
\def\arraystretch{2}
\begin{tabularx}{\textwidth}{ c || *{5}{>{\centering\arraybackslash}X}}
\hline
& DNN classifier (SGD) & DNN classifier (SGD + CycleLR) & DNN classifier  (Adam) & SVM & Proposed model 128x100-AL  \\
\hline
Flower102 &  \makecell{45\\ \footnotesize{($\pm$ 8.5)}} &  \makecell{18\\ \footnotesize{($\pm$ 3.8)}} &  \makecell{32\\ \footnotesize{($\pm$ 5.9)}} &  \makecell{87\\ \footnotesize{($\pm$ 5.3)}} & \textbf{1} \\
\hline
Indoor67 &  \makecell{20\\ \footnotesize{($\pm$ 3.1)}} &  \makecell{9\\ \footnotesize{($\pm$ 1.6)}} &  \makecell{18\\ \footnotesize{($\pm$ 3.2)}} &  \makecell{127\\ \footnotesize{($\pm$ 8.2)}} &\textbf{1} \\
\hline
CUB200 &  \makecell{24\\ \footnotesize{($\pm$ 3.6)}} &  \makecell{9\\ \footnotesize{($\pm$ 1.15)}} &  \makecell{16\\ \footnotesize{($\pm$ 2.2)}} &  \makecell{87\\ \footnotesize{($\pm$ 3.1)}} &\textbf{1} \\
\hline
Dog120 &  \makecell{18\\ \footnotesize{($\pm$ 4.1)}} &  \makecell{9\\ \footnotesize{($\pm$ 0.5)}} &  \makecell{15\\ \footnotesize{($\pm$ 2.2)}} &  \makecell{214\\ \footnotesize{($\pm$ 12.4)}} &\textbf{1} \\
\hline
Stanford40 &  \makecell{12\\ \footnotesize{($\pm$ 2.2)}} &  \makecell{8\\ \footnotesize{($\pm$ 0.63)}} &  \makecell{14\\ \footnotesize{($\pm$ 2.4)}} &  \makecell{79\\ \footnotesize{($\pm$ 5.4)}} &\textbf{1} \\
\hline
\end{tabularx}}
\end{table}

\begin{table}[ht!]
  \caption{Comparison of the training time of 128x100-AL with different classifiers on different domain specific datasets when learning the last class of the dataset after having learned the previous classes. Training time duration is expressed relatively to the one of the proposed model: the lower is the value, the better it is. Values are averaged over 30 runs.}
  \label{speedCompIncr}
  \centering
\resizebox{\columnwidth}{!}{%
\def\arraystretch{2}
\begin{tabularx}{\textwidth}{ c ||*{5}{>{\centering\arraybackslash}X}}
\hline
& DNN classifier (SGD) & DNN classifier (SGD + CycleLR) & DNN classifier  (Adam) & SVM & Proposed model 128x100-AL  \\
\hline

Flower102 &  \makecell{5410\\ \footnotesize{($\pm$ 977)}} & \makecell{2222 \\ \footnotesize{($\pm$ 428)}} & \makecell{3867\\ \footnotesize{($\pm$ 655)}} & \makecell{10460\\ \footnotesize{($\pm$ 247)}} & \textbf{1} \\
\hline
Indoor67 & \makecell{1602\\ \footnotesize{($\pm$ 248)}} & \makecell{762\\ \footnotesize{($\pm$ 120)}} & \makecell{1474\\ \footnotesize{($\pm$ 238)}} & \makecell{10207\\ \footnotesize{($\pm$ 101)}} &\textbf{1} \\
\hline
CUB200 & \makecell{4632\\ \footnotesize{($\pm$ 836)}} & \makecell{1836\\ \footnotesize{($\pm$ 271)}} & \makecell{3197\\ \footnotesize{($\pm$ 411)}} & \makecell{17111\\ \footnotesize{($\pm$ 1028)}} &\textbf{1} \\
\hline
Dog120 & \makecell{2703\\ \footnotesize{($\pm$ 582)}} & \makecell{1333\\ \footnotesize{($\pm$ 22)}} & \makecell{2222\\ \footnotesize{($\pm$ 337)}} & \makecell{31239\\ \footnotesize{($\pm$ 376)}} &\textbf{1} \\
\hline
Stanford40 & \makecell{667\\ \footnotesize{($\pm$ 111)}} & \makecell{409\\ \footnotesize{($\pm$ 16)}} & \makecell{737\\ \footnotesize{($\pm$ 126)}} & \makecell{4220\\ \footnotesize{($\pm$ 44)}} &\textbf{1} \\
\hline

\end{tabularx}}
\end{table}

Table~\ref{transferComp} shows that the proposed model is slightly inferior to the others methods in terms of performance on the tested datasets except for the CUB200 dataset. However, due to the learning algorithm used, it is considerably faster to train as shown in Table~\ref{speedComp}. In order to moderate these results, it may be possible to decrease a bit the training time of the other models by using a more aggressive early stop function and different hyper-parameters but the order of magnitude will remain similar: generally the proposed model is 10 times faster to train than the others when trained on the whole dataset at once. However, as explained above, if it is required to only learn a new class, both the deep neural network classifier and the SVM have to be retrained from scratch and thus, the proposed model is exceptionally faster to train, as shown in Table~\ref{speedCompIncr}.

\section{Conclusion}
We introduced a new classifier model primarily composed of Self-Organizing Maps and Sparse Associative Memories. By combining the proposed layer with a pre-trained Deep Neural Network, it is possible to design flexible deep neural networks with a considerably faster learning algorithm as well as the support of incremental learning at the cost of a slight decrease of the accuracy. Using the proposed model can help the development and the deployment of new intelligent embedded devices which can learn new elements in real time and adapt themselves to their environment without the help of an external agent responsible for the execution of the training algorithm.

\end{document}